\documentclass{article}

\usepackage{arxiv}

\usepackage[utf8]{inputenc} 
\usepackage[T1]{fontenc}    
\usepackage{hyperref}       
\usepackage{url}            
\usepackage{booktabs}       
\usepackage{amsfonts}       
\usepackage{nicefrac}       
\usepackage{microtype}      
\usepackage{lipsum}
\usepackage{graphicx}
\usepackage{subfigure}
\usepackage{booktabs}
\usepackage{amsmath}
\usepackage{amssymb}
\usepackage{mathtools}
\usepackage{amsthm}
\usepackage{diagbox}
\usepackage{authblk}
\usepackage[capitalize,noabbrev]{cleveref}
\graphicspath{ {./images/} }

\title{Revisiting QMIX: Discriminative Credit Assignment by Gradient Entropy Regularization}

\author[1]{\textbf{Jian Zhao$^*$}}
\author[1]{\textbf{Yue Zhang$^*$}}
\author[1]{\textbf{Xunhan Hu}}
\author[2]{\textbf{Weixun Wang}}
\author[1]{\textbf{Wengang Zhou\\}}
\author[2]{\textbf{Jianye Hao}}
\author[3]{\textbf{Jiangcheng Zhu}}
\author[1]{\textbf{Houqiang Li}}
\affil[1]{University of Science and Technology of China}
\affil[2]{College of Intelligence and Computing, Tianjin University}
\affil[3]{Huawei Cloud}

\begin{document}
\maketitle
\def\thefootnote{*}\footnotetext{These authors contributed equally to this work}
\begin{abstract}
In cooperative multi-agent systems, agents jointly take actions and receive a team reward instead of individual rewards.
In the absence of individual reward signals, credit assignment mechanisms are usually introduced to discriminate the contributions of different agents so as to achieve effective cooperation.
Recently, the value decomposition paradigm has been widely
adopted to realize credit assignment, and QMIX has become the state-of-the-art solution.
In this paper, we revisit QMIX from two aspects.
First, we propose a new perspective on credit assignment measurement and empirically show that QMIX suffers limited discriminability on the assignment of credits to agents.
Second, we propose a gradient entropy regularization with QMIX to realize a discriminative credit assignment, thereby improving the overall performance.
The experiments demonstrate that our approach can comparatively improve learning efficiency and achieve better performance.
\end{abstract}


\section{Introduction}
\label{introduction}
Cooperative multi-agent reinforcement learning has been widely adopted in various domains, such as autonomous cars~\cite{cao2012overview}, sensor networks~\cite{zhang2011coordinated}, and robot swarms~\cite{huttenrauch2017guided,busoniu2008a}.
In these tasks, due to the unavailability of individual rewards, each agent is required to learn a decentralized policy through a shared team reward signal~\cite{sunehag2017value,hernandez-leal2019a}.
Thereby, discriminative credit assignment among agents plays an essential role in achieving effective cooperation.
Recent years have witnessed great advances of cooperative multi-agent reinforcement learning methods in credit assignment. 
Among them, value-based methods have shown state-of-the-art performance on challenging tasks, \emph{e.g.}, unit micromanagement in StarCraft II~\cite{samvelyan19smac}.

Recently, value factorization has become particularly popular, which is based on the centralized training with decentralized execution (CTDE) paradigm ~\cite{ LandonKraemer2016MultiagentRL}.
Specifically, it factorizes joint value function $Q_{tot}$ by the integration of individual value functions $Q_i$ during centralized training~\cite{sunehag2017value}.
During the execution, decentralized policies can be easily derived by greedily selecting individual actions from the local value function $Q_i$.
In this way, an implicit multi-agent credit assignment is realized because $Q_i$ is learned by optimizing the total temporal-difference error on the single global reward signal.

As the first attempt of CTDE, VDN~\cite{sunehag2017value} factorizes $Q_{tot}$ by leveraging summation operation over individual $Q_i$~\cite{StuartRussell2003QdecompositionFR}.
Despite its effectiveness in realizing the scalability of multi-agent joint policy training and individual policy execution, VDN simply assigns equal credits to agents, which gives insufficient incentive for agents to learn cooperation.
Considering the limit of equal assignment, QMIX designs the mixing network to assign the non-negative learnable weights to individual Q-values with a non-linear function of the global state~\cite{rashid2018qmix}.
Following QMIX, more and more algorithms employ the variants of attention mechanism as the integration function~\cite{yang2020qatten,wang2020qplex}.

However, an open question is whether the mixing networks really achieve discriminative credit assignment.
In this work, towards quantitative evaluation on such discriminability, we employ the gradient entropy of $Q_{tot}$ over $Q_i$ to measure the discrepancy among the credits assigned to agents.
The empirical results of the QMIX on gridworld environments and the benchmark environment SMAC show that the gradient entropy is always close to the maximum value during training, implying the limited credit assignment in QMIX.
Besides, we simplify the QMIX network by using the summation of $Q_i$ as the input of mixing network instead of $\{Q_i\}$, denoted as QMIX-simple.
In this way, the gradients of $Q_{tot}$ over $Q_i$ are equal.
The experiments show that QMIX-simple can achieve comparable performance as QMIX, which also supports the above argument.

Based on the above observations, we improve QMIX by introducing the gradient entropy regularization.
Since more indiscriminate credit assignment means larger gradient entropy regularization value, gradient entropy regularization can explicitly help the system achieve differentiated credit assignment.
The experimental results demonstrate that gradient entropy regularization not only speeds up the training process but also improves the overall performance.


To sum up, the main contributions of this paper lie in three aspects:
\begin{itemize}
    \item We propose a metric, \emph{i.e.}, the gradient entropy of $Q_{tot}$ over $\{Q_i\}$, to measure the discriminativeness of credit assignment among agents.
    \item We empirically show that QMIX suffers indiscriminability on the credit assignment: (1) the gradient entropy is close to the maximum value, \emph{i.e.}, all agents are assigned the same credit; (2) the simplified version of QMIX can achieve comparable performance as QMIX.
    \item We employ the gradient entropy regularization to explicitly enhance the discriminality on the credit assignment. The experimental results on the gridworld and the benchmark environment demonstrate the efficacy of our method.
\end{itemize}

\section{Related Work}
\label{related work}

In this section, we briefly review the related work on credit assignment in multi-agent reinforcement learning,
which can be divided into Q-learning and policy-gradient.

\subsection{Credit Assignment in Q-learning}
Value decomposition is a popular credit assignment method utilized in algorithms where a centralized Q-value mixer is adopted.
These methods strive to decompose a centralized Q value into individual Q values.
VDN \cite{sunehag2017value} uses a simple sum operation to generate the global Q value, which equally treats the contributions of all agents.
QMIX \cite{rashid2018qmix}, as an extension of VDN,
employs a mixing network by leveraging state information to decide the transformation weights \cite{ha_hypernetworks_2016}, which allows non-linearity operations to decompose the central Q value.
QATTEN \cite{yang2020qatten} utilizes a multi-head attention structure to weight the individual Q-values based on the global state and the individual features, and then linearly integrates these values into the central action value.
QPLEX \cite{wang2020qplex} decomposes the central Q-value into the sum of individual value functions and a non-positive advantage function.
By introducing the duplex dueling structure, QPLEX achieves the complete function class that satisfies the IGM principle.
WQMIX \cite{rashid_weighted_2020} follows QMIX paradigm and exploits a weighted QMIX operator to put more emphasis on better joint actions.

\subsection{Credit Assignment for Policy Gradient}
Existing Credit assignments for policy gradient utilize a similar implementation of value decomposition.
MADDPG \cite{lowe2017multi} first introduces DDPG \cite{silver2014deterministic} method to multi-agent tasks.
A centralized Q-value estimator is used to aggregate agents' information.
It achieves an implicit and weak credit assignment by giving a global view of all agents' actions to centralized critic.
Counterfactual Multi-Agent Policy Gradients (COMA) \cite{foerster2018counterfactual} further gives a counterfactual baseline of expectation of all agents' Q-values.
Since the computation of counterfactual baseline fixes all other agents' actions, it cannot reflect the contribution of each agent.
Essentially, COMA solves the lazy agent problem by encouraging agents to explore while not hurting the effect of the joint policy.
Off-policy multi-agent decomposed policy gradients (DOP) \cite{wang2020off} takes advantage of both QMIX and MADDPG, and investigates how to implement value decomposition in the mulit-agent actor-critic framework.
LICA \cite{zhou_learning_2020} integrates global information into the centralized critic and adds a regularization term to encourage stochasticity in policies.

The above methods all express their implicit realization of credit assignment.
To the best of our knowledge, there is no study on the metric to quantitatively evaluate the discriminability of credits assigned by these methods.
In this work, we propose a reasonable evaluation metric for discriminability of credit assignment, and introduce the gradient entropy regularization to help mixing network realize a more discriminative credit assignment.

\section{Preliminaries}
\label{background}

\subsection{Problem Formulation}
A fully cooperative multi-agent problem is often described as a Dec-POMDP \cite{MartinLPuterman1994MarkovDP,FransAOliehoek2016ACI}, which consists of a tuple $G=<S,U,P,r,Z,O,n,\gamma>$. 
$s\in S$ represents the global state of the environment. 
At each timestep, each agent $a\in A\equiv \{ 1,2,\cdots,n \}$ chooses an action $u^a\in U$, forming a joint action $\pmb{u}\in U^n$. 
Given this joint action, the environment state will transit into a new one according to the transition function $P(s^\prime |s,\pmb{u}):S\times U^n\times S\rightarrow [0,1]$ ~\cite{MichaelLLittman1994MarkovGA, JunlingHu2003NashQF, LucianBusoniu2008ACS}.
The global reward $r(s,\mathbf{u})$ is shared by all the agents and $\gamma\in [0,1)$ is a discount factor.  

Due to the communication constraints in practice, we consider the partially observable setting, where agents have no access to the global state for decision making. 
Each agent could only draw individual observations $z\in Z$ from the environment according to the observation function $O(s,a):S\times A\rightarrow Z$.
Generally, each agent keeps an observation-action history $\tau^a\in T\equiv (Z\times U)^*$ to better derive the global information from limited local observations.
With $\tau^a$, agent $a$ can determine its action with policy $\pi^a(u^a|\tau^a):T\times U\rightarrow [0,1]$.
The individual policies are gathered to be a joint policy $\pi$, which corresponds to a joint action-value function, \emph{i.e.} $Q(s_t,\pmb{u}_t)=\mathbb{E}[R_t|s_t,\pmb{u}_t]$, where $R_t=\sum_{i=0}^{+\infty}\gamma^i r_{t+i}$ represents the discounted reward.
The overall target of the problem is to maximize the value of the joint policy, formulated as $max_{\pi} V(\pi)=\mathbb{E}_{s_0\sim d(s_0),\pmb{u}\sim \pi}[Q(s_0, \pmb{u})]$ where $d(s_0)$ is the initial distribution of the global state.

Under the centralized training and decentralized execution paradigm, the policies of agents are only conditioned on local observation-action histories $\tau^a$ but trained with global information, \emph{e.g.}, $s$ and $\pmb{\tau}$.




\subsection{QMIX}

As a representative value decomposition method, QMIX follows the CTDE paradigm.
The core part of QMIX is the mixing network, which is responsible for credit assignment.
In QMIX, each agent $a$ holds an individual Q-network $Q_a(\tau^a, u^a)$ \cite{CarlosGuestrin2001MultiagentPW}.
The output of individual Q-networks are passed through the mixing network, which implicitly assign the credits to each agent and generates the approximation of the global Q-value $Q_{tot}(\pmb{\tau},\pmb{u},s;\theta)$. 
Specifically, the weights of the mixing network are produced by the separate hypernetworks by taking the state $s$ as input.
Each hypernetwork consists of a single linear layer, followed by an absolute activation function, to ensure that the mixing network weights are non-negative.
$Q_{tot}$ is calculated as follows: 
\begin{equation}
\label{qmix_qtot}
    Q_{tot}(\pmb{\tau}, \pmb{u}, s; \theta)=W_2f_{elu}(W_1\pmb{Q}+b_1)+b_2,
\end{equation}
where $\pmb{Q}=[Q_1(\tau_1, a_1), Q_2(\tau_2, a_2), \cdots, Q_n(\tau_n, a_n)]^T\in \mathbb{R}^{n\times 1}$ is the output of individual Q-networks, and $W_1 \in \mathbb{R}_{+}^{m\times n}$, $W_2 \in \mathbb{R}_{+}^{1 \times m}$, $b_1\in \mathbb{R}^{m \times 1}$ and $b_2\in \mathbb{R}$ are weights produced by hypernetworks.
Since all elements in $W_1$ and $W_2$ are non-negative, QMIX satisfies the following condition:
\begin{equation}
    \label{qmix nonnegative}
    \frac{\partial Q_{tot}}{\partial Q_a} \geq 0, \forall a\in A.
\end{equation}
This property guarantees the Individual-Global Maximum (IGM) principle, which means the optimal individual actions jointly make the optimal joint action for $Q_{tot}$:
\begin{equation}
\label{IGM}
    \mathop{argmax}\limits_{\pmb{u}}Q_{tot}(\pmb{\tau}, \pmb{u}, s)=
    \left(
    \begin{matrix}
        \mathop{argmax}\limits_{u^1}Q_{1}(\tau^1, u^1) \\
        \vdots \\
        \mathop{argmax}\limits_{u^n}Q_{n}(\tau^n, u^n)
    \end{matrix}
    \right).
\end{equation}
In this way, agents could choose the best local actions based only on their own local observation-action histories and the joint action is the best action for the whole system. 

The system is updated by minimizing squared TD loss on $Q_{tot}$, according to the following formulation:
\begin{equation}
\label{qmix update}
    \mathbb{L}(\theta)=\sum_{i=1}^b(y_i^{tot}-Q_{tot}(\pmb{\tau},\pmb{u},s;\theta))^2.
\end{equation}
Here $y_i^{tot}=r+\gamma max_{\pmb{u^\prime}} Q_{tot}(\pmb{\tau^\prime},\pmb{u^\prime},s^\prime;\theta^-)$ is the TD target. 

To sum up, QMIX takes the advantage of global information in decomposing the $Q_{tot}$ and guarantees consistency between the centralised and decentralised policies with monotonic constraint.
However, QMIX does not explicitly require discrimination on credit assignment, which might result in indiscriminability and negatively influence the performance.
\section{Methodology}

\begin{figure*}[htbp]
	\centering
	\subfigure[Lumberjacks]{
		\begin{minipage}{0.25\textwidth}
			\includegraphics[width=\textwidth]{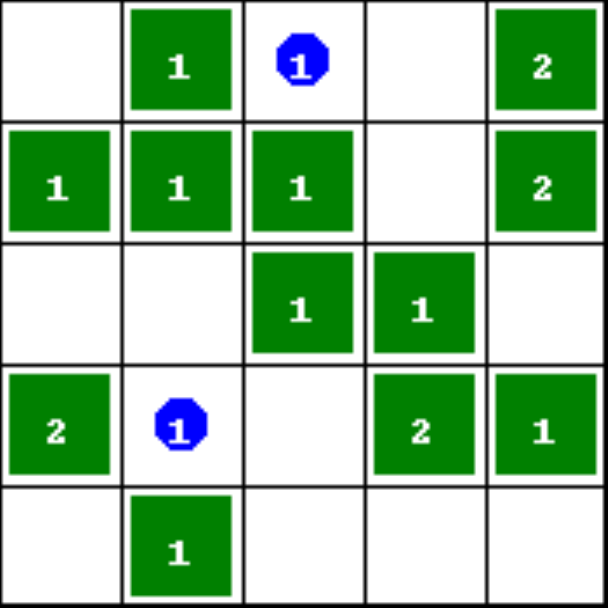}
		\end{minipage}
	}
	\hspace{0.9cm}
	\subfigure[TrafficJunction10]{
		\begin{minipage}{0.25\textwidth} 
            \includegraphics[width=\textwidth]{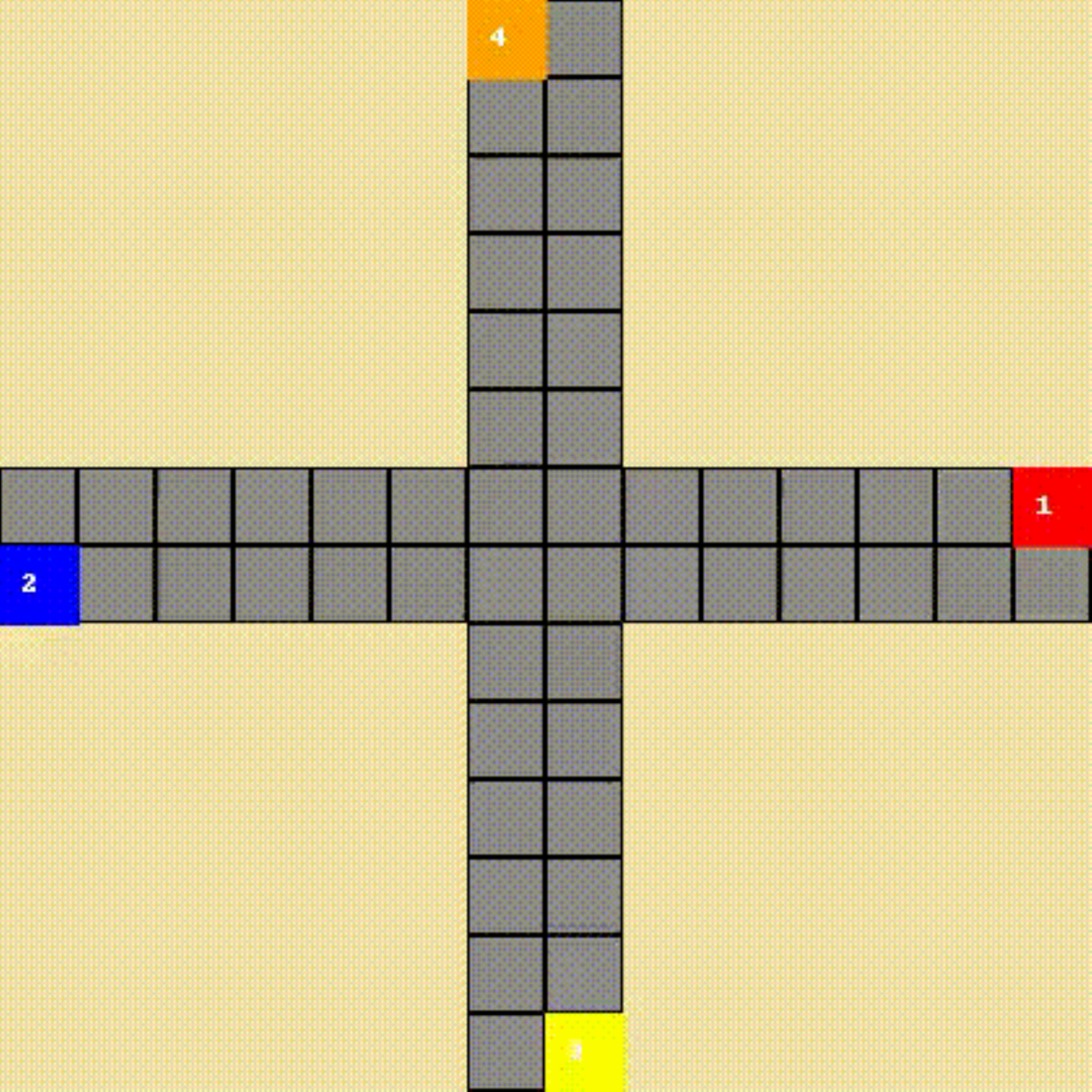} 
		\end{minipage}
	}
	\hspace{0.9cm}
	\subfigure[SMAC]{
		\begin{minipage}{0.25\textwidth}
			\includegraphics[width=\textwidth]{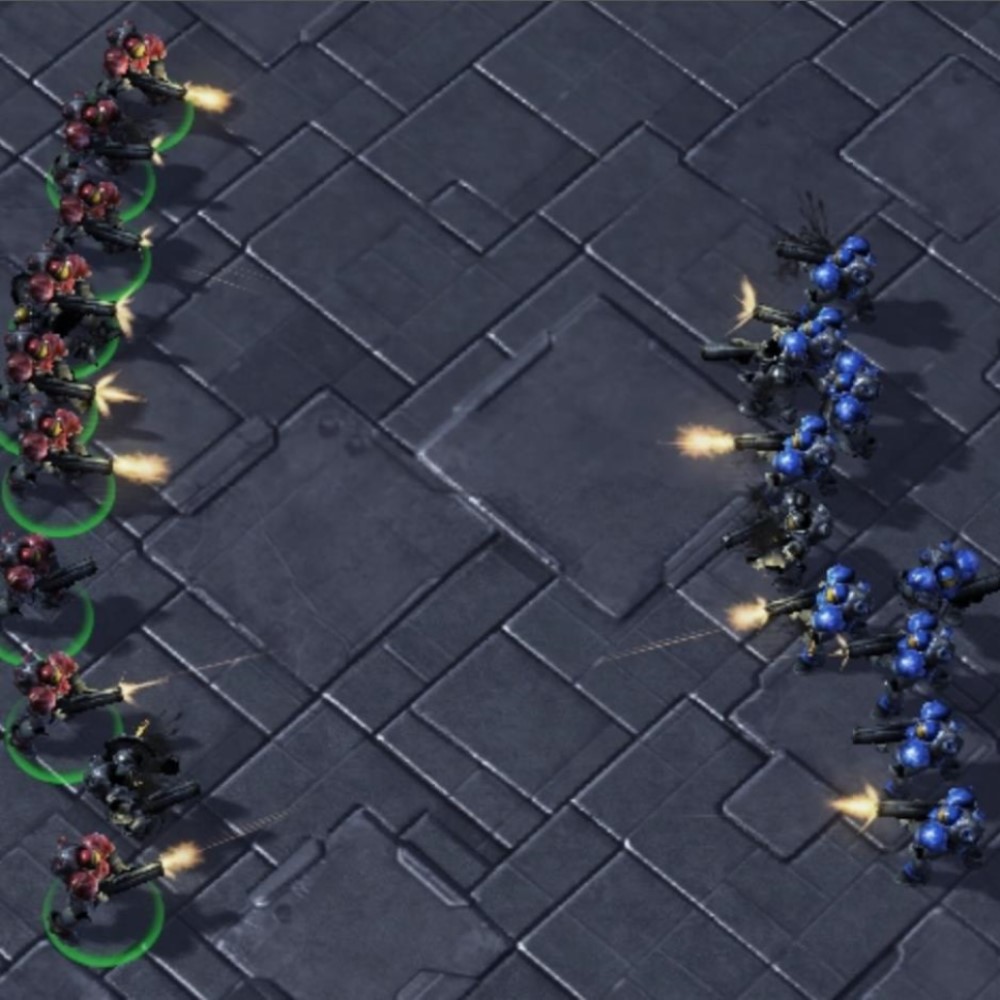}
		\end{minipage}
	}
	\caption{Visualisation of the three experimental environments.} 
	\label{grid_world}
\end{figure*}
In this section, we propose two strategies to investigate the discriminability of credit assignment in mixing network, \emph{i.e.}, an evaluation metric for discriminability measurement and a simplified version of QMIX.
Moreover, we design a gradient entropy regularization to enhance the discriminability of credit assignment.

\subsection{Discriminability Measurement of Credit Assignment via Gradient Entropy}
\label{measurement}
In this section, we propose an evaluation metric to measure the discriminability of credit assignment, and explain the rationale behind it.
Moreover, we employ this metric to evaluate the discriminability of credit assignment in QMIX.

Value decomposition paradigm approximates the global Q-value $Q_{tot}$ by the integration of a set of individual Q-values $Q_i$.
According to Taylor expansion, $Q_{tot}$ can be factorized as follows:
\begin{equation}
    \label{taylor}
    \begin{aligned}
    Q_{tot}&=Q_{tot}^0+(\frac{\partial Q_{tot}}{\partial \pmb{Q}})^T\pmb{Q}+o(\pmb{Q}^T\pmb{Q}) \\
    &\approx Q_{tot}^0+[g_1,\dots,g_n]^T\pmb{Q},
    \end{aligned}
\end{equation}
where $g_i$ denotes the gradient of $Q_{tot}$ over $Q_i$.
The above equation indicates that the influence of each agent on $Q_{tot}$ is determined by the corresponding gradient.
Moreover, a larger gradient means that the corresponding agent has a greater impact on $Q_{tot}$.

Considering that the gradient reflects each agent's contribution to the team, if the credits assigned to agents are discriminative, the gradient distribution over agents is unlikely to be uniform. To describe such non-uniformity, since the gradients are non-negative, we normalize them by Eq.~\eqref{normal} and adopt the entropy of the normalized gradients as a measure of the discriminability of credits assigned to agents.
\begin{equation}
    \label{normal}
    g_i^\prime=\frac{g_i}{\sum_{j=1}^n g_j}
\end{equation}
Beyond that, considering different complexity across tasks, we also normalize the entropy with the maximum entropy of $n$ random variables.
Based on the above discussion, the normalized gradient entropy is calculated as follows:
\begin{equation}
    \label{entropy}
    H_{norm}(\frac{\partial Q_{tot}}{\partial \pmb{Q}})=(-\sum_{i=1}^n g_i^\prime \log g_i^\prime)/\log n,
\end{equation}
where $\log n$ is the maximum entropy of $n$ values.
It can be seen that the smaller the entropy, the more discriminative the gradient, and vice versa.
In other words, if the normalized gradient entropy is close to 1, the gradients of $Q_{tot}$ on $\{ Q_i \}$ will be similar, meaning a lack of discriminability in credit assignment.

Given the definition of normalized gradient entropy, we employ it to measure the discriminability of credits assigned by QMIX.
It is worth noting that the normalized gradient entropy is applicable to measure the discriminability of any differentiable mixing network structure.
According to the mixing network structure in QMIX, the gradients of $Q_{tot}$ over $\{Q_i\}$ are calculated as follows:
\begin{equation}
    \label{gradient}
    [g_1,\dots,g_n]^T = \frac{\partial Q_{tot}}{\partial \pmb{Q}}=W_1^T F W_2,
\end{equation}
where $F=diag\{ \frac{df_{elu}(x)}{dx}|_{x=x_1}, \cdots, \frac{df_{elu}(x)}{dx}|_{x=x_n} \}$ is the gradient on the activation function $f_{elu}$. 
$x_i$ here is the $i$-th element in the output of the first linear layer in mixing network. 
As shown in Eq.~\eqref{qmix nonnegative}, each element of the gradient vector is non-negative.
The detailed derivation process can be found in supplementary material.

\subsection{Simplified QMIX}
\label{qsimple}

\begin{figure}[htb]
	\centering
	\subfigure{
		\begin{minipage}{0.55\textwidth}
          \includegraphics[width=\textwidth]{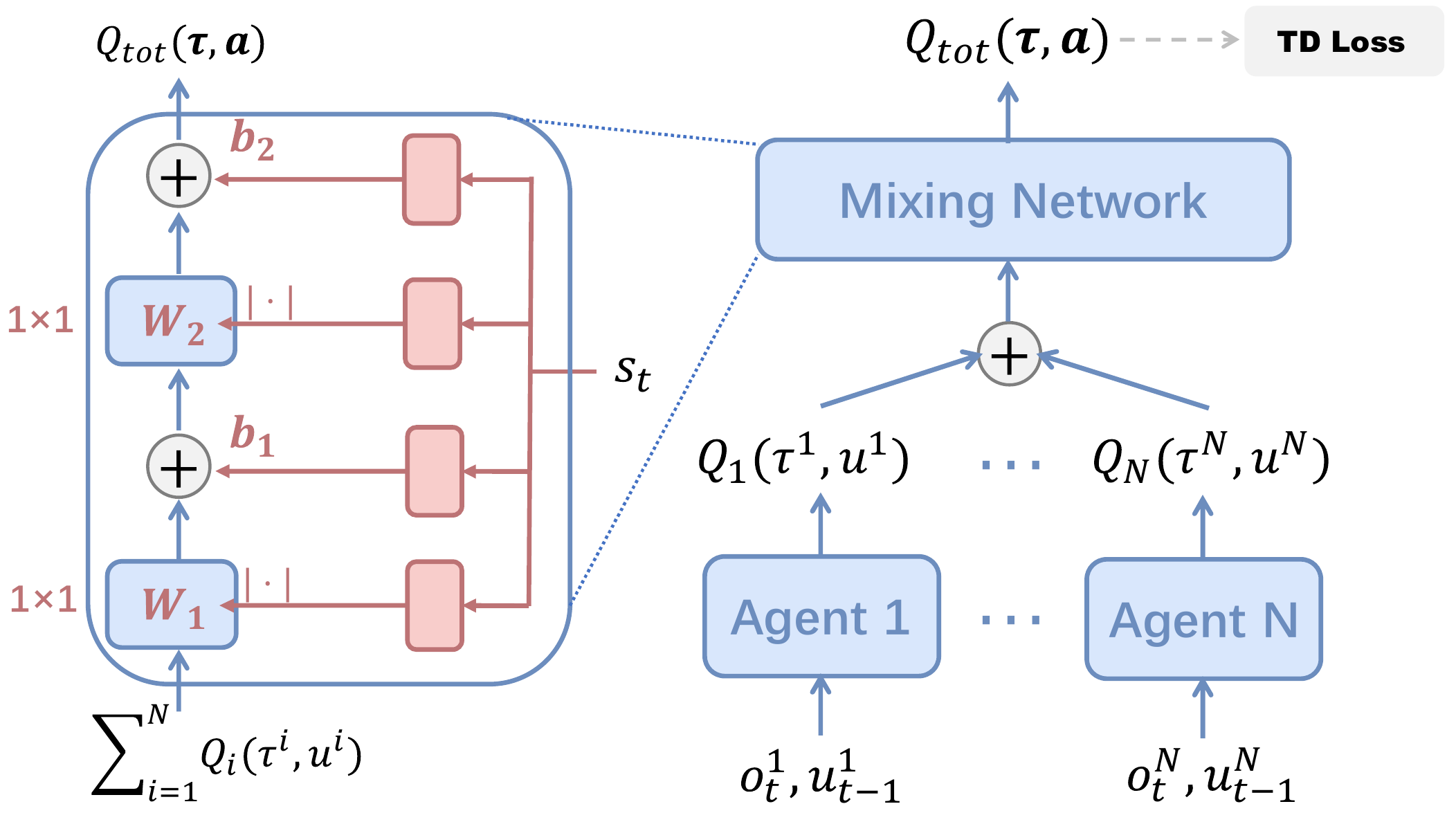} 
		\end{minipage}
	}
	\caption{Simplified QMIX framework.
	Different from original QMIX, the simplified QMIX directly take the summation of individual $Q_i$s rather than $Q_i$ vector as the input of the mixing network.
	The parameters $W_1$ and $W_2$ produced by hypernetwork are scalars, compared to matrix in QMIX.} 
	\label{qsimple_framework}
\end{figure}

\begin{figure*}[htb]
    
	\centering
	\subfigure[Lumberjacks]{
	    \label{qmix_entropy_curve_Lumber}
		\begin{minipage}{0.3\textwidth}
            \includegraphics[width=\textwidth]{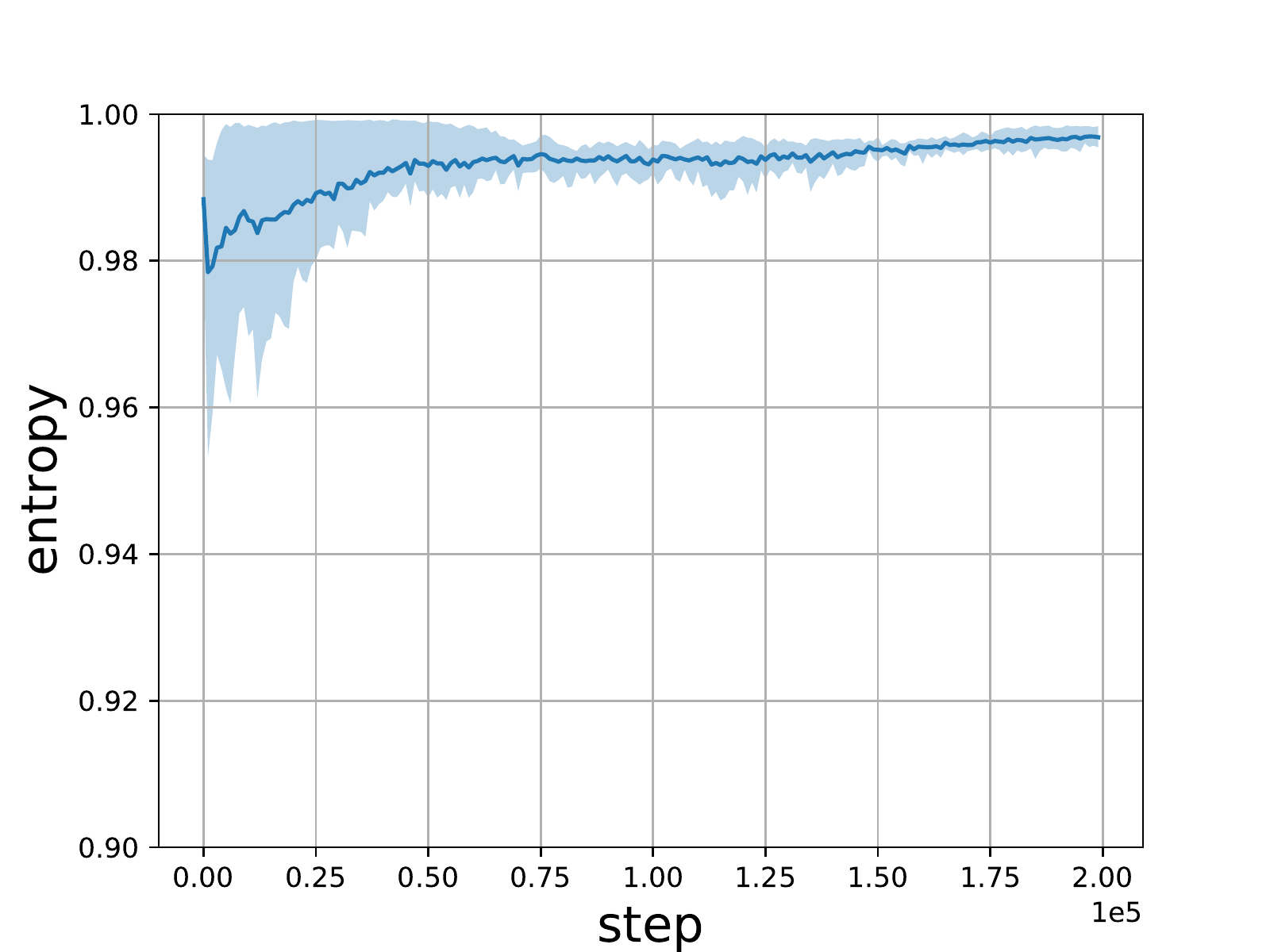}
		\end{minipage}
	}
	\subfigure[TrafficJunction10]{
	    \label{qmix_entropy_curve_Traffic10}
		\begin{minipage}{0.3\textwidth}
            \includegraphics[width=\textwidth]{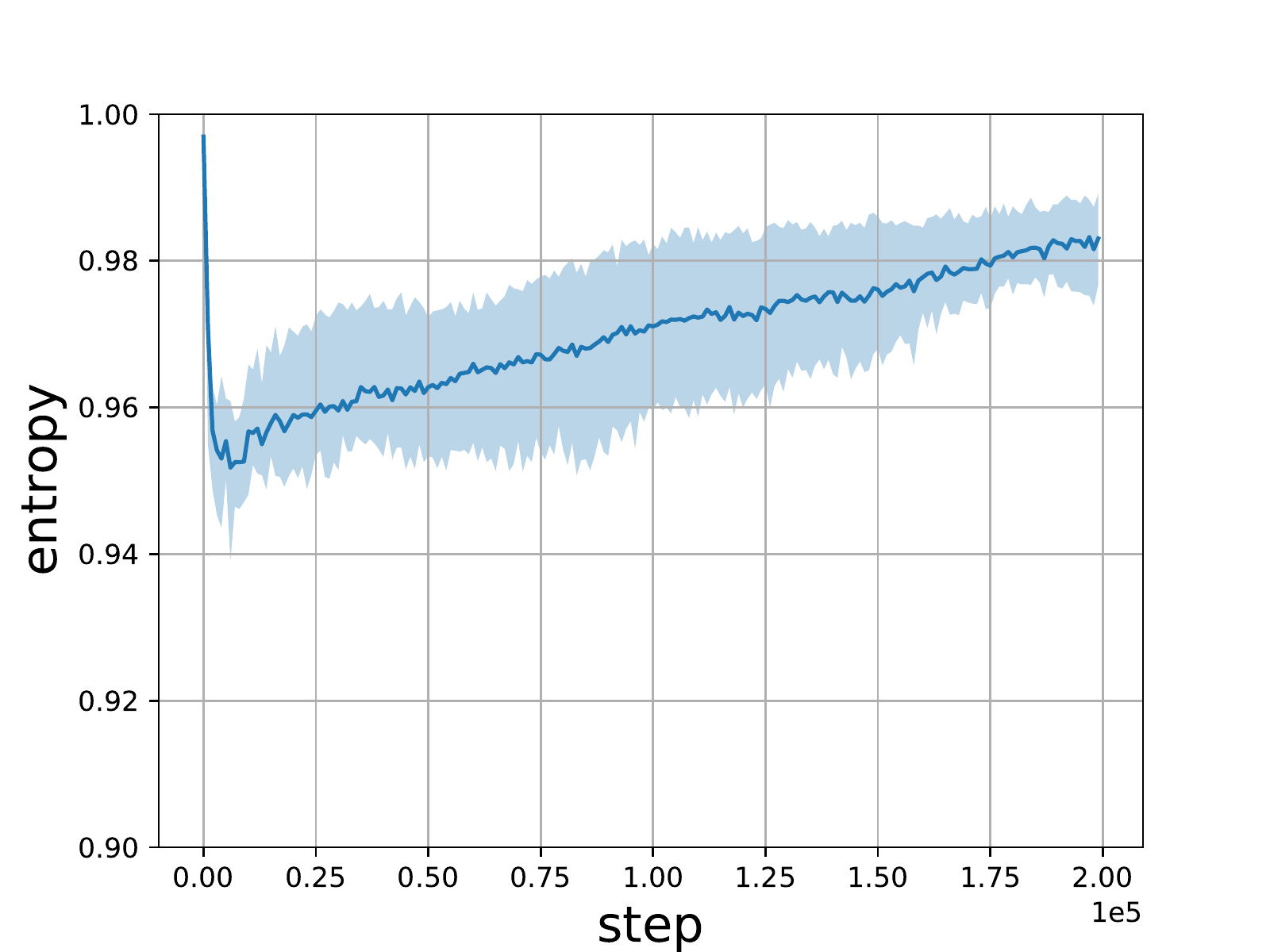}
		\end{minipage}
	}
	\\
	\resizebox{0.9\linewidth}{!}{\subfigure[3s5z]{
	    \label{qmix_entropy_curve_3s_vs_5z}
		\begin{minipage}{0.3\textwidth}
            \includegraphics[width=\textwidth]{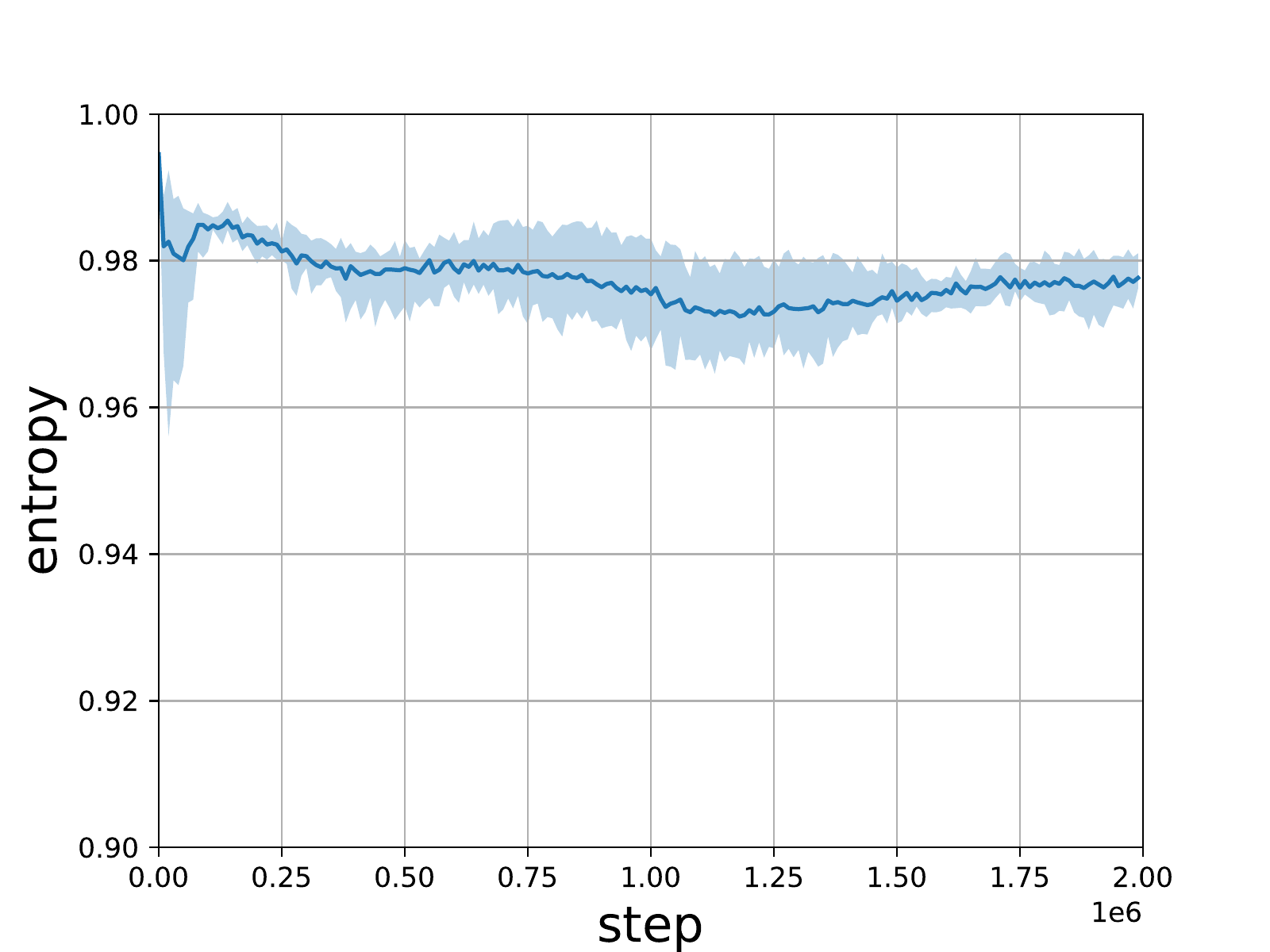}
		\end{minipage}
	}
	\subfigure[MMM2]{
	    \label{qmix_entropy_curve_3s5z}
		\begin{minipage}{0.3\textwidth}
            \includegraphics[width=\textwidth]{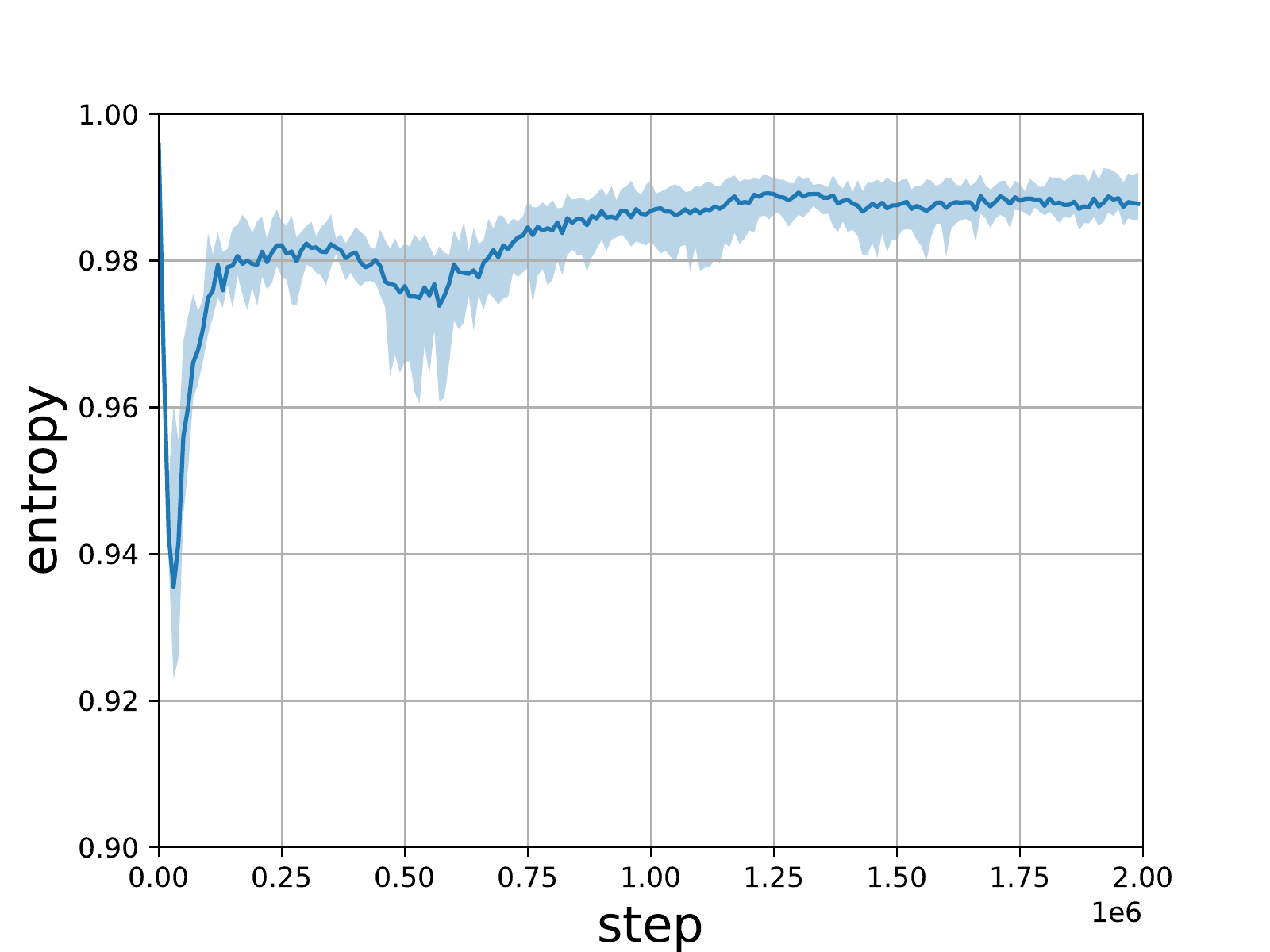}
		\end{minipage}
	}
	\subfigure[27m vs 30m]{
	    \label{qmix_entropy_curve_27m_vs_30m}
		\begin{minipage}{0.3\textwidth}
            \includegraphics[width=\textwidth]{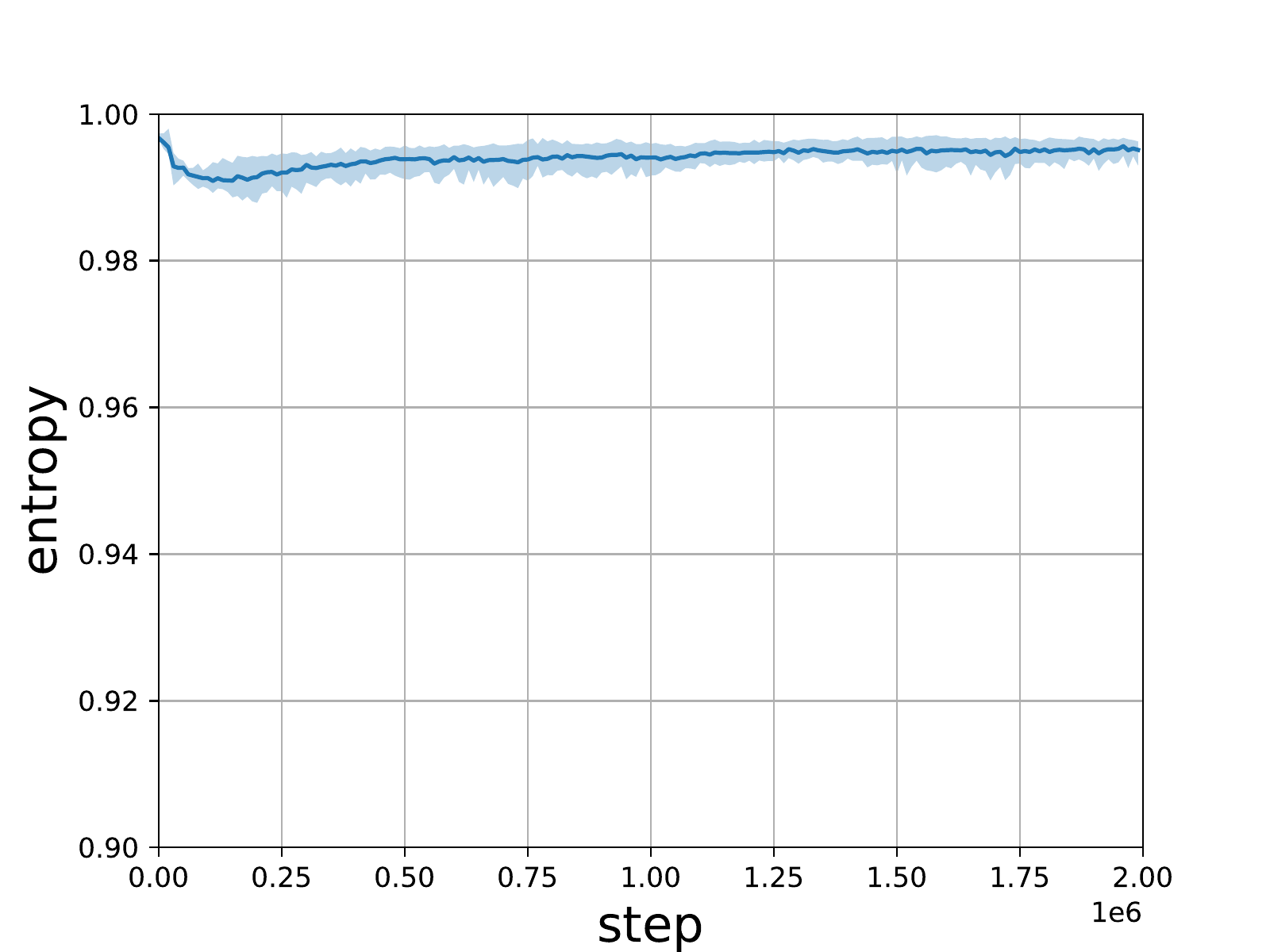}
		\end{minipage}
	}}
	\caption{Entropy curve during QMIX training (with y-axis ranging from 0.9 to 1.0).
	In (a) \emph{Lumberjacks} and (b) \emph{TrafficJunction10}, the data is collected as batch mean of normalized entropy every 1k steps.
	In (c) \emph{3s5z}, (d) \emph{MMM2} and (e) \emph{27m vs 30m} from SMAC, the data is collected every 10k steps.
	The solid line here shows the mean of the normalized entropy on five different random seeds.
	The upper and lower bound of shadow area represents the min and max entropy among 5 different random seeds.
	In these environments, the entropy keeps close to the maximum value (1.0) during the whole process except at the random exploration part, which means that the mixing network tends to assign equal credits to all the agents.} 
	\label{qmix_entropy_curve}
\end{figure*}
In addition to the discriminability measurement via gradient entropy, we conduct the ablation study to test the discriminability of QMIX in credit assignment.
To this end, we propose a simplified version of QMIX, named QMIX-simple, as a baseline.

QMIX-simple follows a similar paradigm as QMIX but simplifies the mixing strategy.
The overall framework of QMIX-simple is shown in Figure~\ref{qsimple_framework}.
Different from QMIX that integrates individual Q-values via learnable non-negative weights, QMIX-simple applies the summation operation over individual Q-values, followed by two scalar multiplications with an activation function in between.
Specifically, $Q_{tot}$ of QMIX-simple, denoted as $Q_{tot}^{simple}$, is calculated with the following equation:
\begin{equation}
    \label{qmix simple}
    Q_{tot}^{simple}=w_2 f_{elu}(w_1 (\sum_{i=1}^n Q_i)+b_1)+b_2,
\end{equation}
where $w_1 \in \mathbb{R}_+$, $w_2 \in \mathbb{R}_+$, $b_1 \in \mathbb{R}$ and $b_2 \in \mathbb{R}$ are produced by hypernetworks that take the global state as input.
In this way, the gradients on $Q_i$ are forced to be equal and credits are assigned equally.

\begin{table}[htbp]
    \centering
    \resizebox{0.6\linewidth}{!} {
    \begin{tabular}{c|c|c}
        \hline
        Model & Equal Credits & Utilization of Global State \\
        \hline
        \textbf{VDN} & $\surd$ & $\times$ \\ 
        \textbf{QMIX} & $\times$ & $\surd$ \\ 
        \textbf{QMIX-simple} & $\surd$ & $\surd$ \\ 
        \hline
    \end{tabular}
    }
    \caption{Comparisons of three models.}
    \label{tab:model_diff}
\end{table}

\textbf{Discussion:} 
Here, we give a brief discussion of QMIX-simple, QMIX and VDN.
Table~\ref{tab:model_diff} summarizes the comparison of these three models.
VDN takes the sum of individual Q-values to approximate the global Q-value, which also assigns equal weights to agents.
The key difference between VDN and QMIX-simple is that QMIX-simple employs scalar multiplications and non-linear activation function to enrich the expressiveness of mixing network.
Besides, QMIX-simple takes the advantage of global state to get a state-dependent multiplicator while VDN actually ignores the global state.
Compared to QMIX, QMIX-simple also utilizes the global state in mixing network but explicitly assigns equal weights.
If QMIX-simple could achieve comparable performance as QMIX, it means that QMIX lacks discriminability in credit assignment.

\subsection{Gradient Entropy Regularization}
In this section, we propose a new method to guide QMIX to allocate different credits to different agents, named Gradient entropy REgularization (GRE).

In Section~\ref{measurement}, we have explained the rationality of employing gradient entropy to measure the discriminability of credit assignment.
Thereby, one straightforward way to enhance the discriminability of credit assignment is to incorporate the gradient entropy regularization into QMIX.
The overall loss of QMIX-GRE is the combination of the original TD loss and the regularizatioin term, \emph{i.e},
\begin{equation}
    \begin{aligned}
    \mathbb{L}&=\mathbb{L}_{TD_{loss}} + \lambda\mathbb{L}_{Reg}\\
    &=\sum_{i=1}^b[(y_i^{tot}-Q_{tot}(\pmb{\tau},\pmb{u},s;\theta))^2+ \lambda H_{norm}(\frac{\partial Q_{tot}}{\partial \pmb{Q}})],
    \end{aligned}
    \label{reg}
\end{equation}
where $\lambda$ is a hyperparameter balancing the TD loss and the regularization term.
Note that when $\lambda = 0$, the model degenerates to the original QMIX.
A larger value of $\lambda$ means greater penalty for indiscriminability.
It is worth mentioning that optimizing the regularization term only updates the parameters in the mixing network.

\begin{table*}[htbp]
\centering
\resizebox{0.7\textwidth}{!}{
\begin{tabular}{c|ccccc}
\hline
\diagbox{Percentile}{Env} & Lumberjacks & TrafficJunction10 & 3s5z & MMM2  & 27m vs 30m \\ \hline
5\%        & 0.972       & 0.931             & 0.941    & 0.961 & 0.987      \\
25\%       & 0.992       & 0.961             & 0.975    & 0.980 & 0.992      \\
50\%       & 0.997       & 0.975             & 0.984    & 0.988 & 0.995      \\
75\%       & 0.999       & 0.985             & 0.989    & 0.993 & 0.997      \\
95\%       & 1.000       & 0.993             & 0.995    & 0.996 & 0.999     \\ \hline
\end{tabular}}
\caption{The percentiles of gradient entropy of QMIX.
The data is collected during the whole training process and sorted from smallest to largest.
$i^{th}$ percentile value represents the $i\%$ smallest value in the data.
In all our experimental environments, the $25^{th}$ percentile of gradient entropy value is close to 1.
Since the maximum entropy means equal probability, the results here reveals the limited discriminablity of credit assignment in QMIX.}
\label{qmix_entropy_quantile}
\end{table*}

In our training process, the neural networks will minimize the TD loss as well as the normalized gradient entropy of the mixing network, forcing itself to assign different credits to different agents in an explicit way.
As a result, the mixing network will learn a more discriminative assignment mechanism.

\section{Experiments}
To comprehensively understand the credit assignment in QMIX and evaluate our proposed gradient entropy regularization, we conduct the experiments to answer the following questions in Section 5.2, 5.3, and 5.4, respectively:

\textbf{RQ1}: Does QMIX achieve a discriminative credit assignment via mixing network?

\textbf{RQ2}: With the help of gradient entropy regularization, how does QMIX-GRE perform compared to QMIX?

\textbf{RQ3}: Can QMIX-GRE enhance the discriminability of credit assignment?

\subsection{Experimental Settings}
In this section, we briefly introduce the experimental environments and the hyperparameter settings.

In this paper, we conduct experiments on two gridworld environments and SMAC environments.
The two gridworld environments, named Lumberjacks and TrafficJunction10, come from ~\cite{magym}.
SMAC is introduced by ~\cite{samvelyan19smac}

\textbf{Lumberjacks:}
The agents are lumberjacks whose goal is to cut down all the trees in the map. 
Each tree has a specified strength when an episode starts. 
For a tree in the map, once the number of agents in the cell of the tree is the same as or greater than its strength, the tree is automatically cut down. 
If all of the trees are cut down or the limit of the maximum episode step is reached, current episode ends.
The possible actions of agents are moving in four directions (Down, Left, Upper, Right) or do nothing.

\textbf{TrafficJunction10:}
The map is a 14 × 14 grid intersection and in each episode, there are 10 vehicles in total waiting to get into this intersection.
At each time step, ``new" cars enter the intersection from each of the four directions with probability $p$ and is randomly assigned to one of three possible routes (go straight, turn left and turn right).
Once a car reaches the edge of the grid, it will be removed. 
The total number of cars in the map at same time is limited. 
Cars are requested to keep on the right side of the road.
The car has two possible actions: move forward to its target or stay at the current cell. 
If the positions of two cars overlap, they collide and get a team reward -10, but this will not affect the simulation.
The reward of any time step is the sum of passed time steps of all the cars in the map with a negative factor and the collide reward.

\textbf{SMAC:}
This environment is built on Blizzard's StarCraft II RTS game and is very popular in MARL research field.
There are many maps in the environment where two teams of different agents try their best to defeat each other.
In the game, agents have no access to the global state and the observations are strictly limited.
In different maps of SMAC, the number and type of agents are different.
The actions of agents also depend on the scenario, whose number ranges from 7 to 70.
Once all the agents in a team die, the other team wins the game.
Although the overall target is to have the highest win rate for each scenario, the environment still offers a reward signal to help training.
This reward signal consists of hit-point damage dealt and received by agents, the units killed and the battle result.

Our implementation is based on PyMARL~\cite{samvelyan19smac} codebase and follows its hyperparameter settings. 
All the experiments in this paper are repeated five times with different random seeds.
The models are trained with 2M steps and tested every 10K steps in SMAC while trained with 200K steps and tested every 5K steps in other two environments.
Our codes are available in \url{https://github.com/sgzZ123/GRE}.

\begin{figure*}[htb]
	\centering
	\subfigure[Lumberjacks]{
		\begin{minipage}{0.25\textwidth}
			\includegraphics[width=\textwidth]{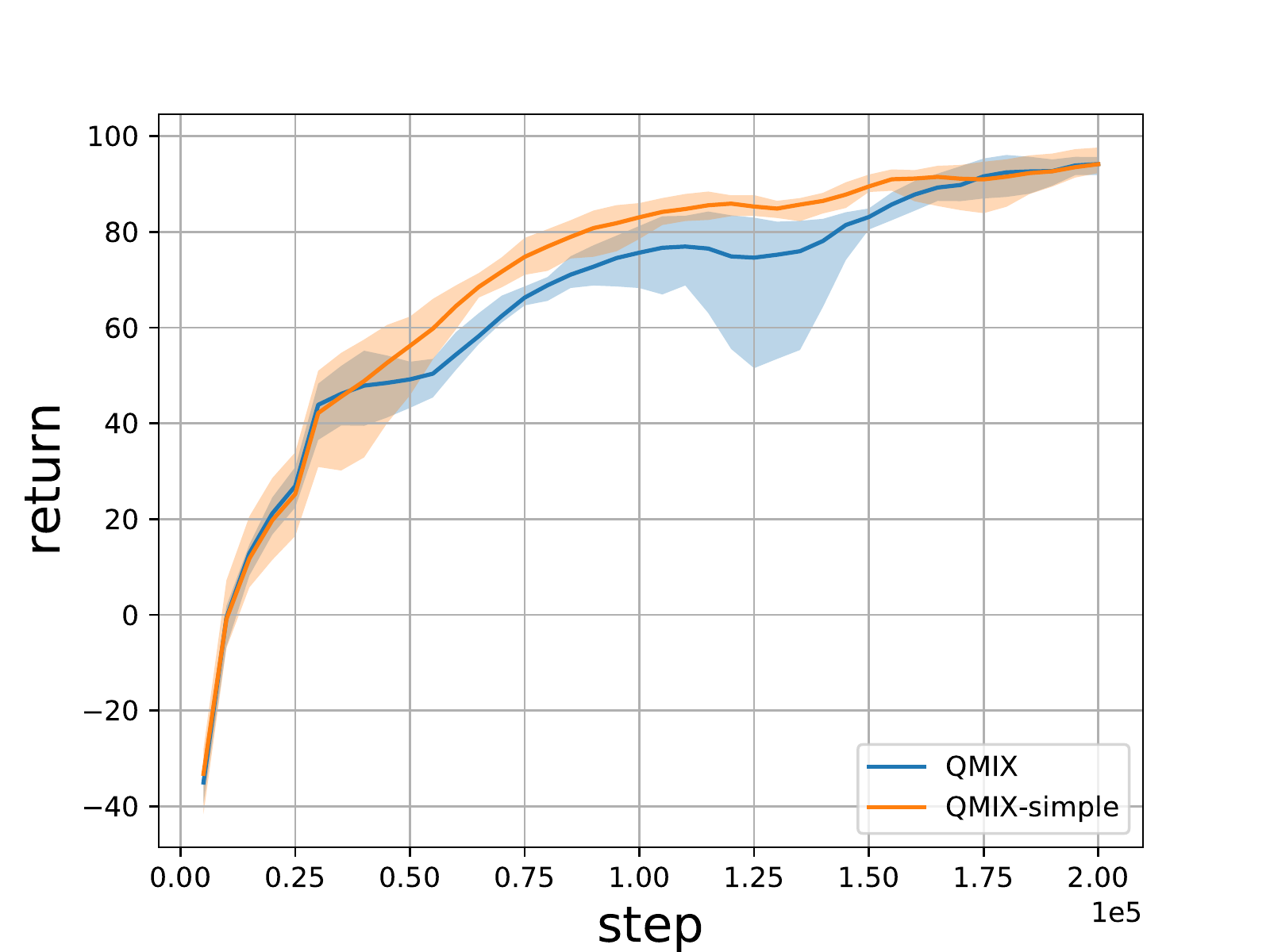}
		\end{minipage}
	}
	\subfigure[TrafficJunction10]{
		\begin{minipage}{0.25\textwidth}
			\includegraphics[width=\textwidth]{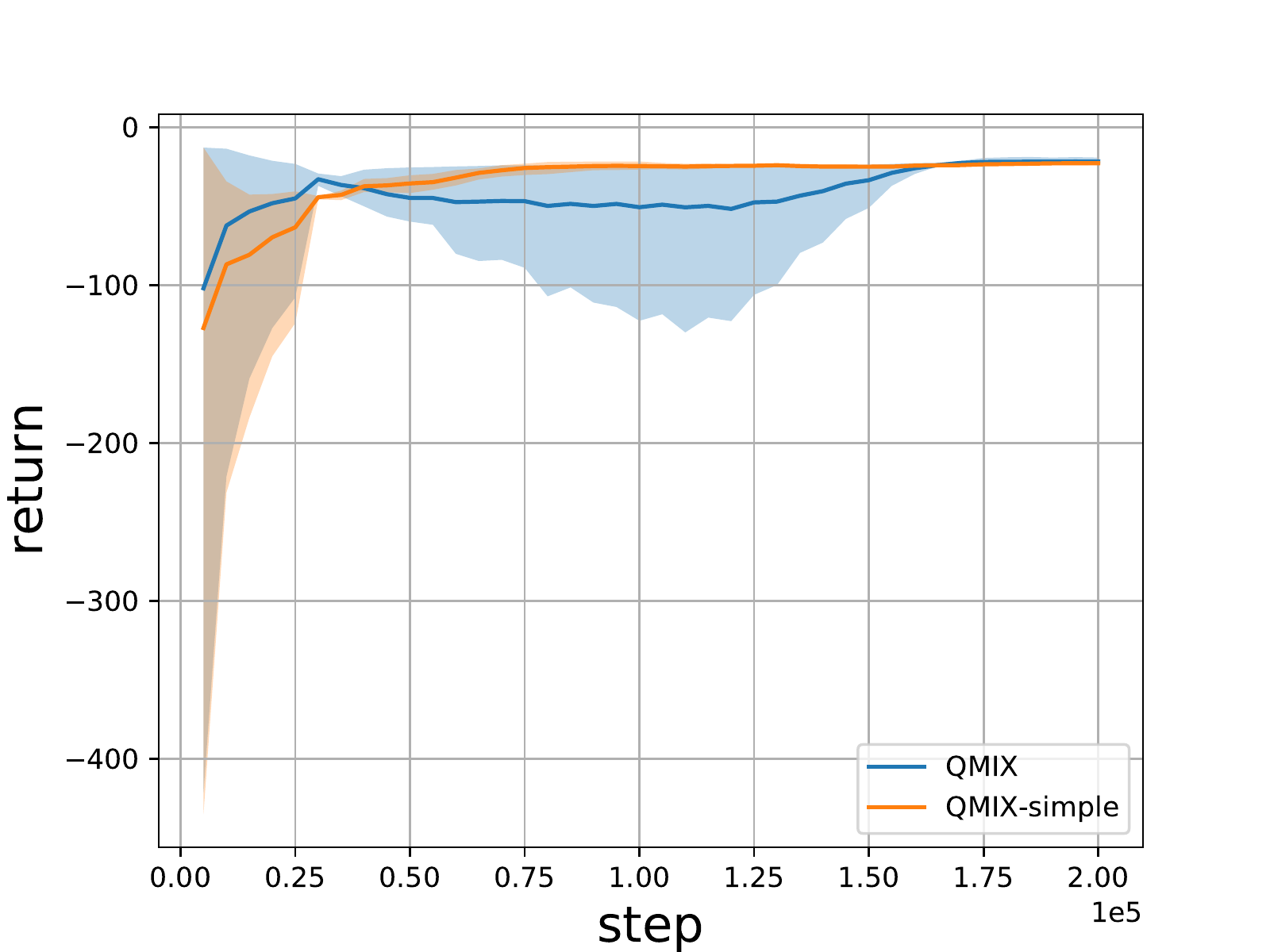}
		\end{minipage}
	}
	\\
	\resizebox{0.8\linewidth}{!}{
	\subfigure[3s5z]{
		\begin{minipage}{0.3\textwidth}
			\includegraphics[width=\textwidth]{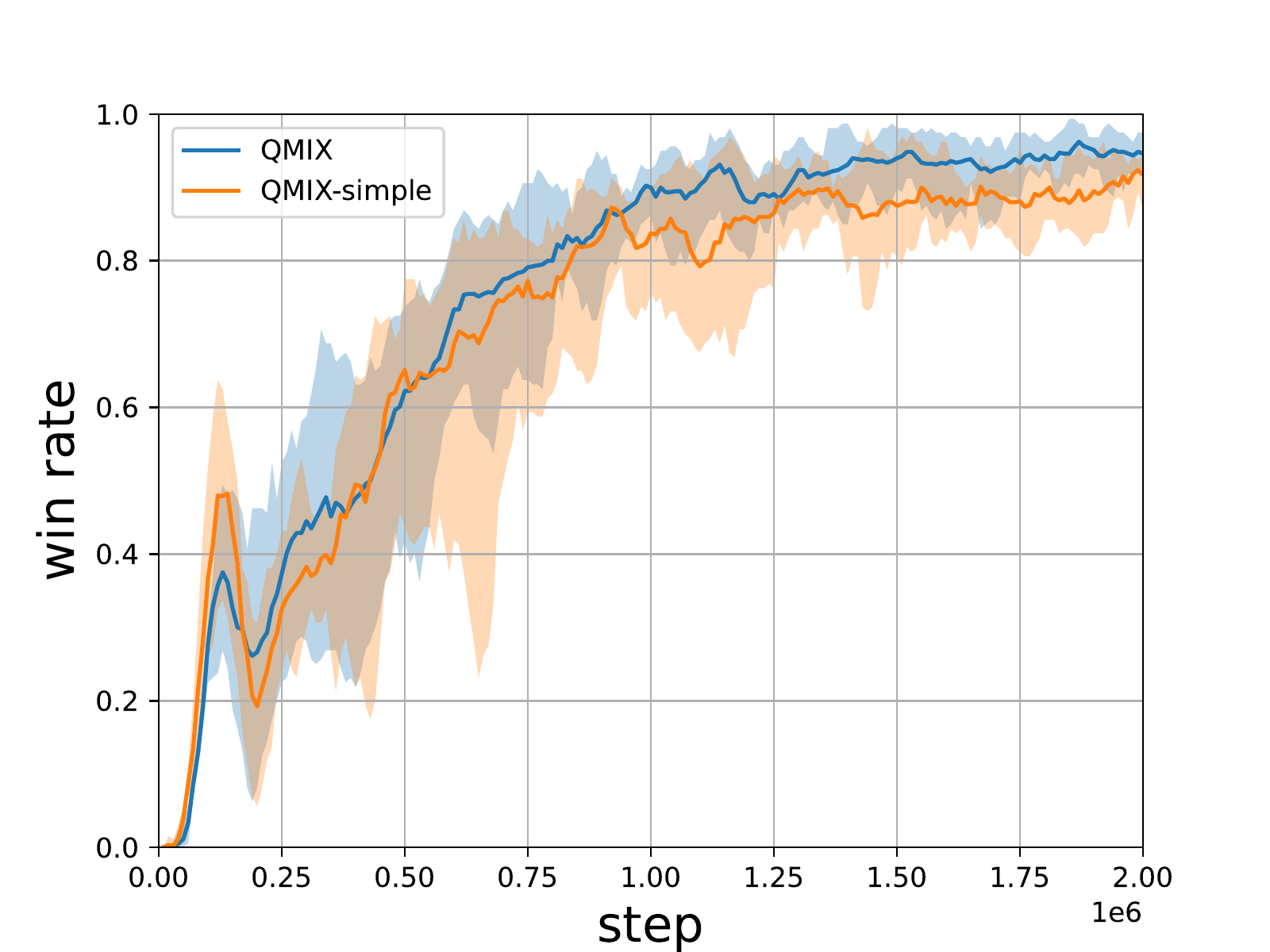}
		\end{minipage}
	}
	\subfigure[MMM2]{
		\begin{minipage}{0.3\textwidth} 
            \includegraphics[width=\textwidth]{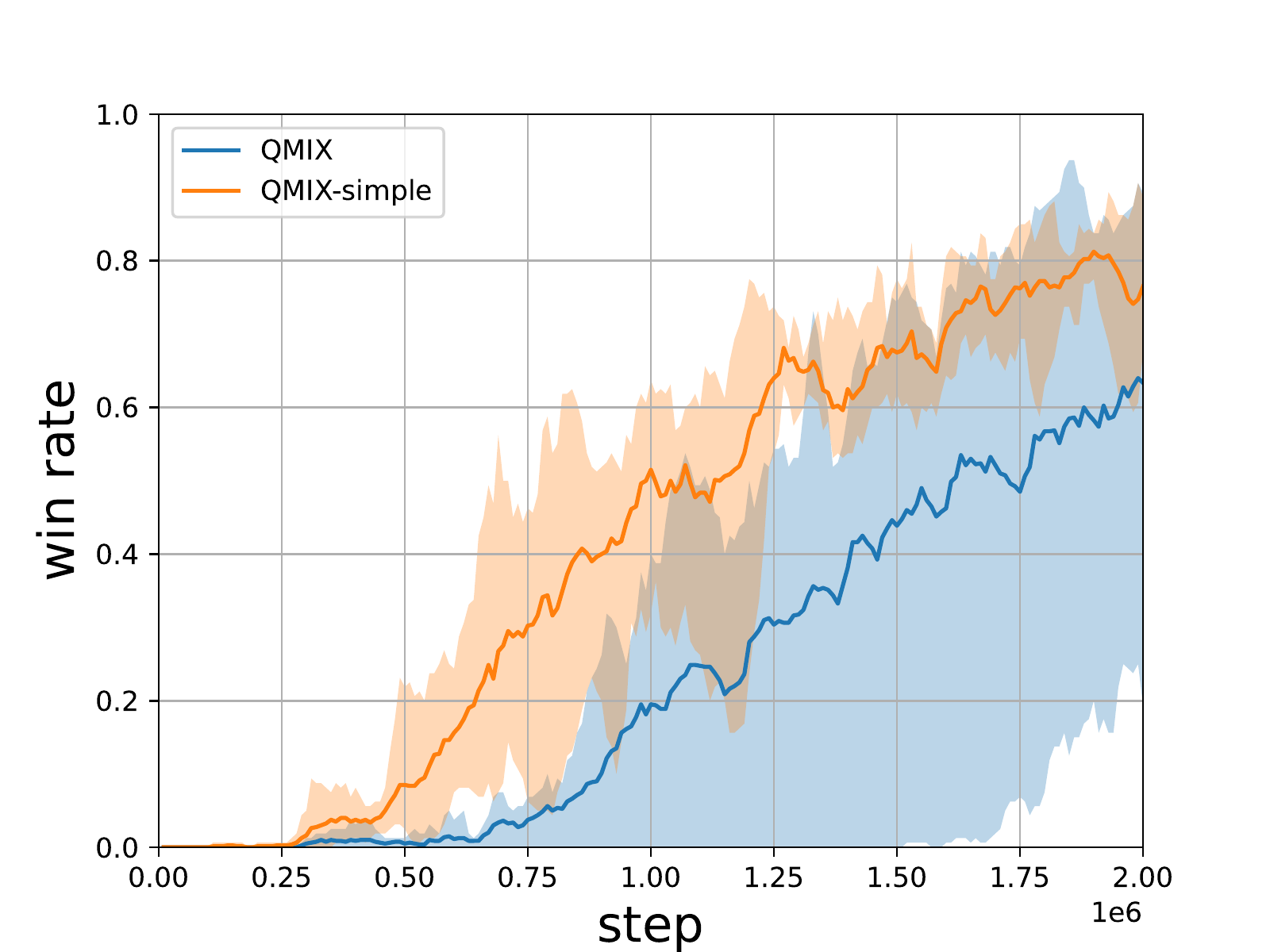} 
		\end{minipage}
	}
	\subfigure[27m vs 30m]{
		\begin{minipage}{0.3\textwidth}
			\includegraphics[width=\textwidth]{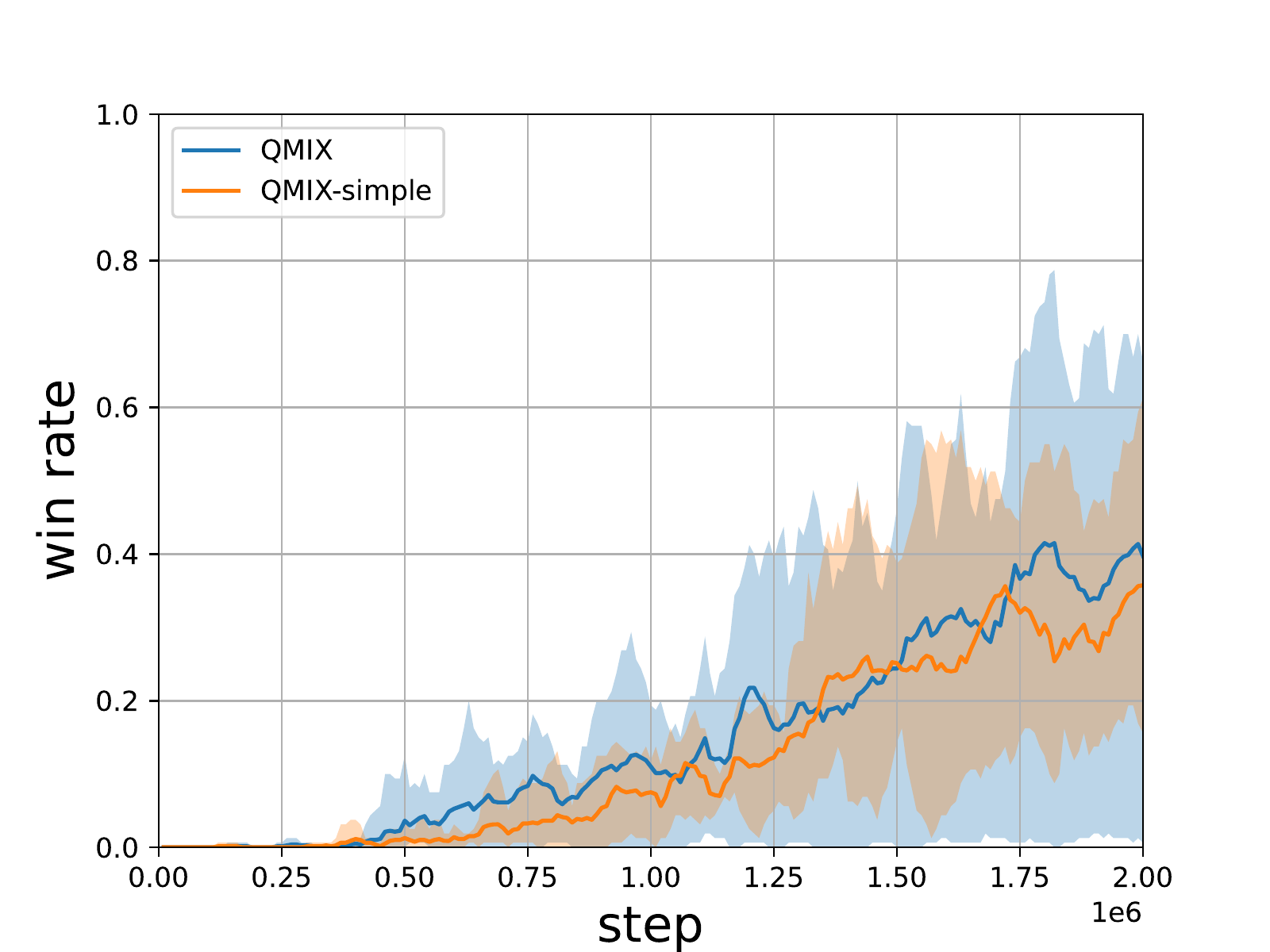} 
		\end{minipage}
	}}
	\caption{The performance of QMIX and QMIX-simple.
	Both of the solid line are the mean performance from 5 different random seeds.
	The upper and lower bound of the shadow area are min and max values in the 5 seeds.
	(a) Lumberjacks and (b) TrafficJunction10 are experiments on gridworld, where QMIX-simple performs the same as QMIX.
	(c) \emph{3s5z}, (d) \emph{MMM2} and (e) \emph{27m vs 30m} are experiments on SMAC.
	In (c), the performance of QMIX-simple is close to the performance of QMIX, although with a tiny drop which probably comes from the decrease of parameters.
	In (d), QMIX-simple even outperforms the original QMIX.
	In (e) QMIX-simple performs the same as original QMIX.
	} 
	\label{qsimple_performance}
\end{figure*}
\subsection{Investigation of QMIX: RQ1}

To evaluate the discriminability of credits assigned by QMIX, we employ the normalized gradient entropy metric and compare the performance of QMIX and QMIX-simple.

To calculate the normalized gradient entropy, we collect the normalized entropy data on various SMAC environments and gridworld environments.
Table~\ref{qmix_entropy_quantile} summarizes the $q^{th}$ ($q \in \{5,25,50,75,95\}$) percentile of entropy values for each map/environment, respectively.
Across the five experimental environments, the differences between the $95^{th}$ and $5^{th}$ percentile of entropy values are small, which indicate that the gradient entropy in QMIX consistently follows a long-tail distribution.
Figure~\ref{qmix_entropy_curve} shows the normalized entropy curve in training process, where the y-axis ranges from 0.9 to 1.0.
We can find the normalized entropy keeps close to the maximal value except for random exploration stage where agents randomly select their actions.
To sum up, the distribution of gradient entropy demonstrates that QMIX suffers discriminability issue on the assignment of credits to agents.

In addition to discriminability measurement via gradient entropy, we conduct the ablation study to test the discriminability of QMIX in credit assignment.
We conduct experiments on various SMAC environments and gridworld environments.
In practice, we change the output of hypernetworks from matrix to scalars and keep other hyperparameters the same as original QMIX.
Due to the page limit, we show the performance of QMIX-simple and QMIX on five representative maps/environments in Figure~\ref{qsimple_performance}.
The results on the remaining five maps of SMAC environments can be found in appendix.
We can find that the performance of QMIX-simple is extremely close to the original QMIX algorithm except for few environments with only a minor drop.
The performance drop probably comes from the decrease in the number of hypernetwork parameters (from matrices to scalars).

In summary, both the large value of gradient entropy and comparable performance of QMIX-simple to QMIX demonstrate that the credit assignment in QMIX still lacks discriminability. 

\subsection{Comparative Results: RQ2}
\begin{figure*}[htbp]
	\centering
	\subfigure[Lumberjacks]{
		\begin{minipage}{0.25\textwidth} 
            \includegraphics[width=\textwidth]{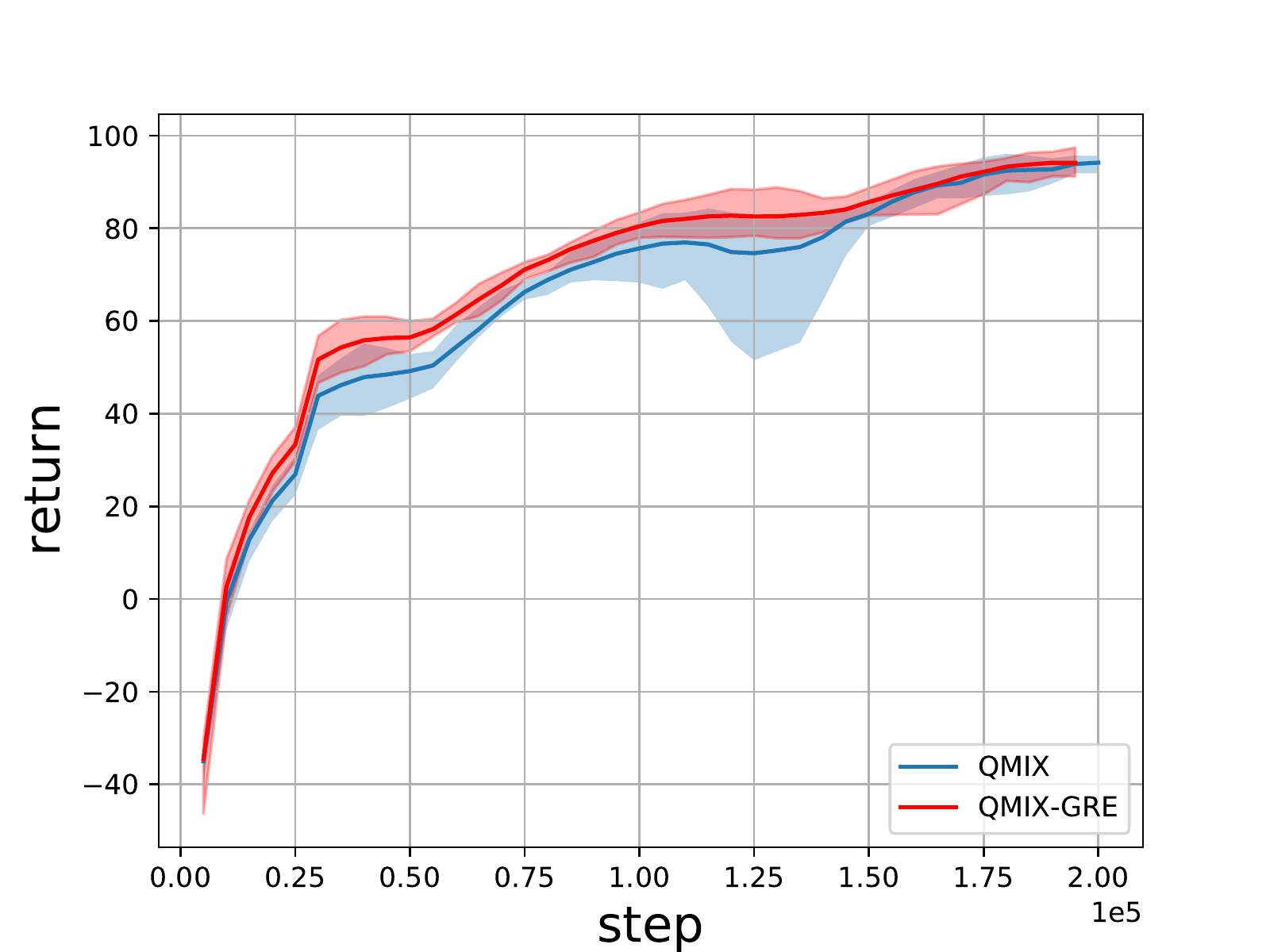}
		\end{minipage}
	}
	\subfigure[TrafficJunction10]{
		\begin{minipage}{0.25\textwidth} 
            \includegraphics[width=\textwidth]{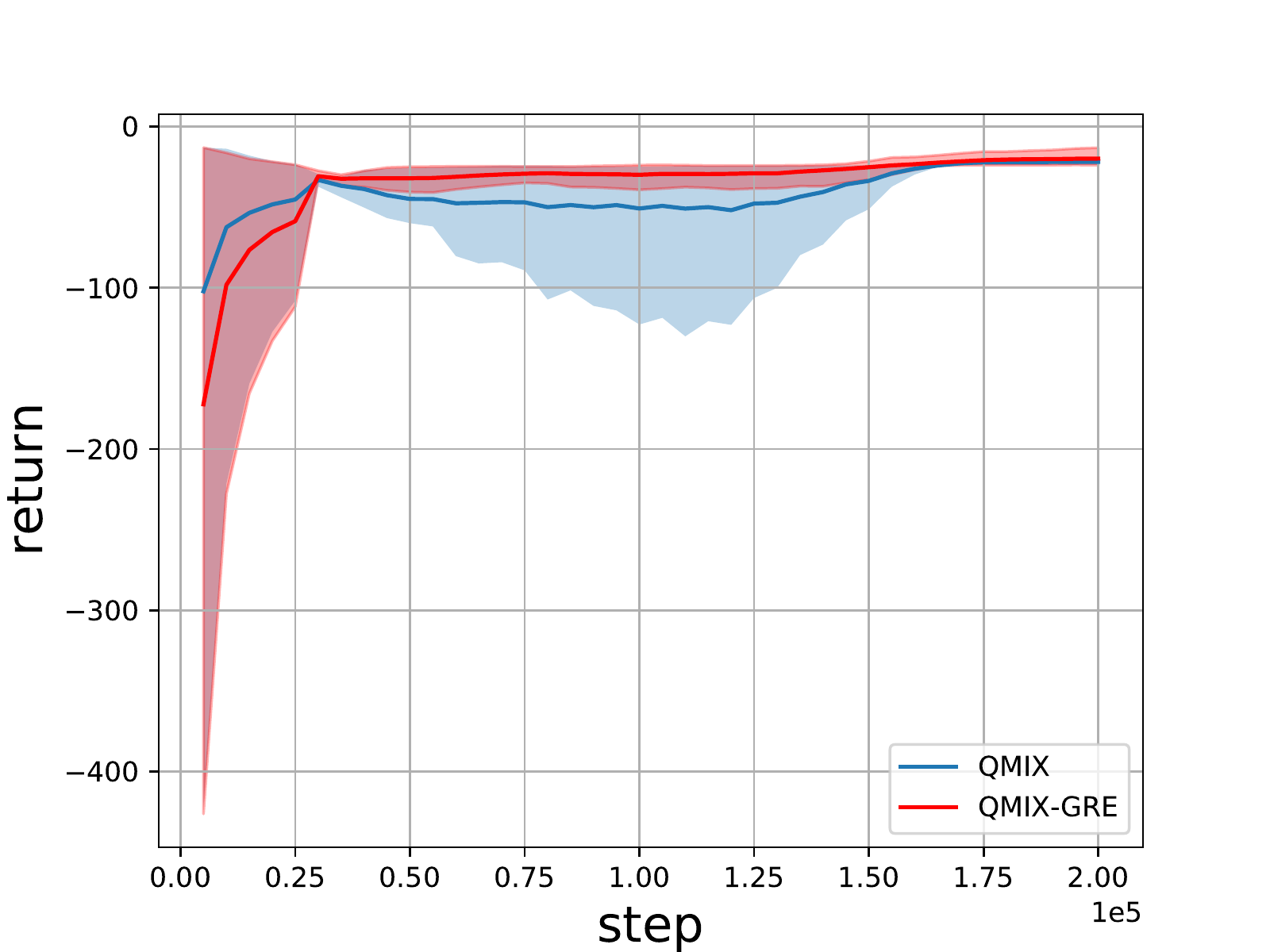}
		\end{minipage}
	}
	\\
	\resizebox{0.8\linewidth}{!}{
	\subfigure[3s5z]{
		\begin{minipage}{0.3\textwidth} 
            \includegraphics[width=\textwidth]{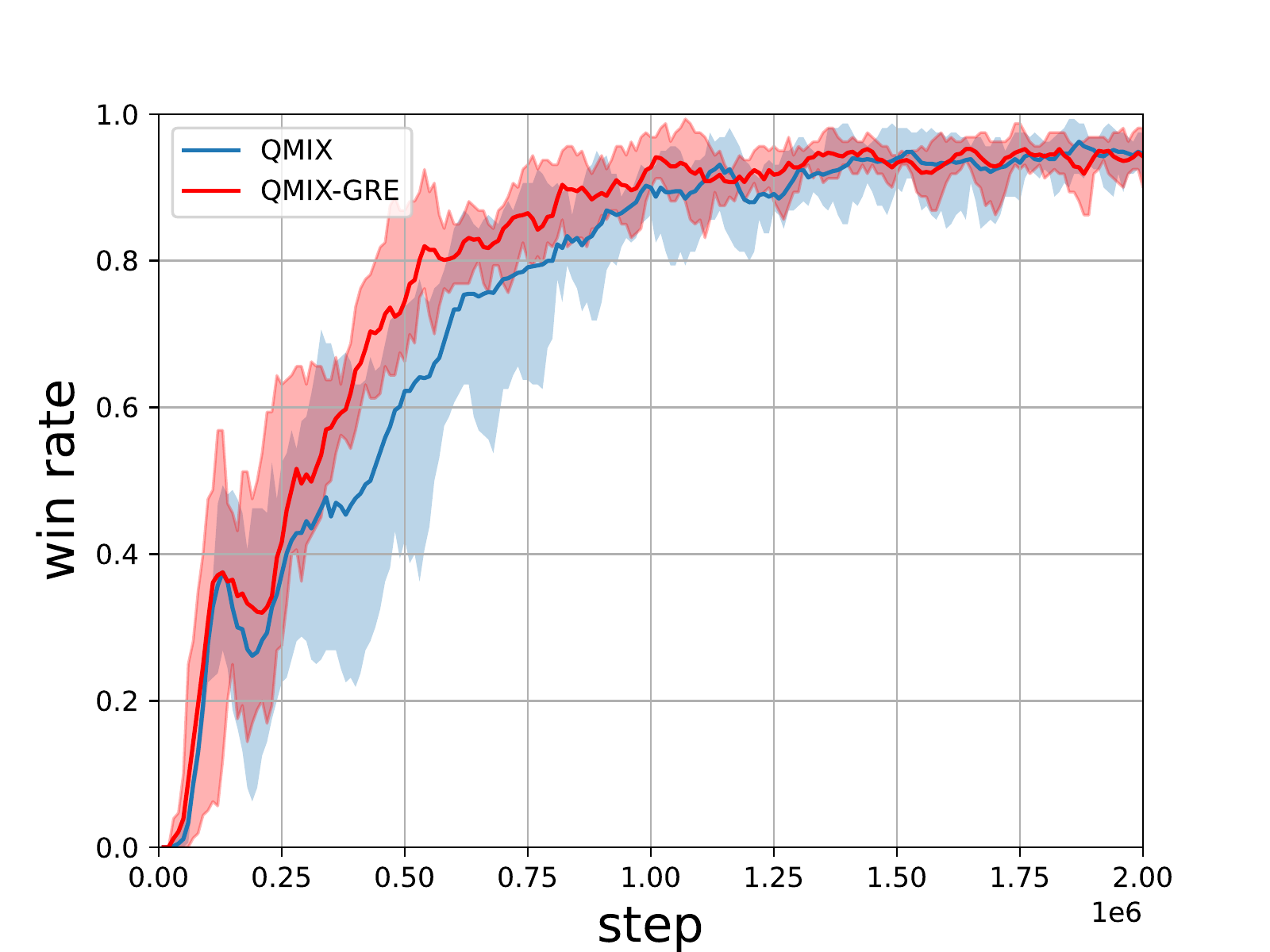} 
		\end{minipage}
	}
	\subfigure[MMM2]{
		\begin{minipage}{0.3\textwidth} 
            \includegraphics[width=\textwidth]{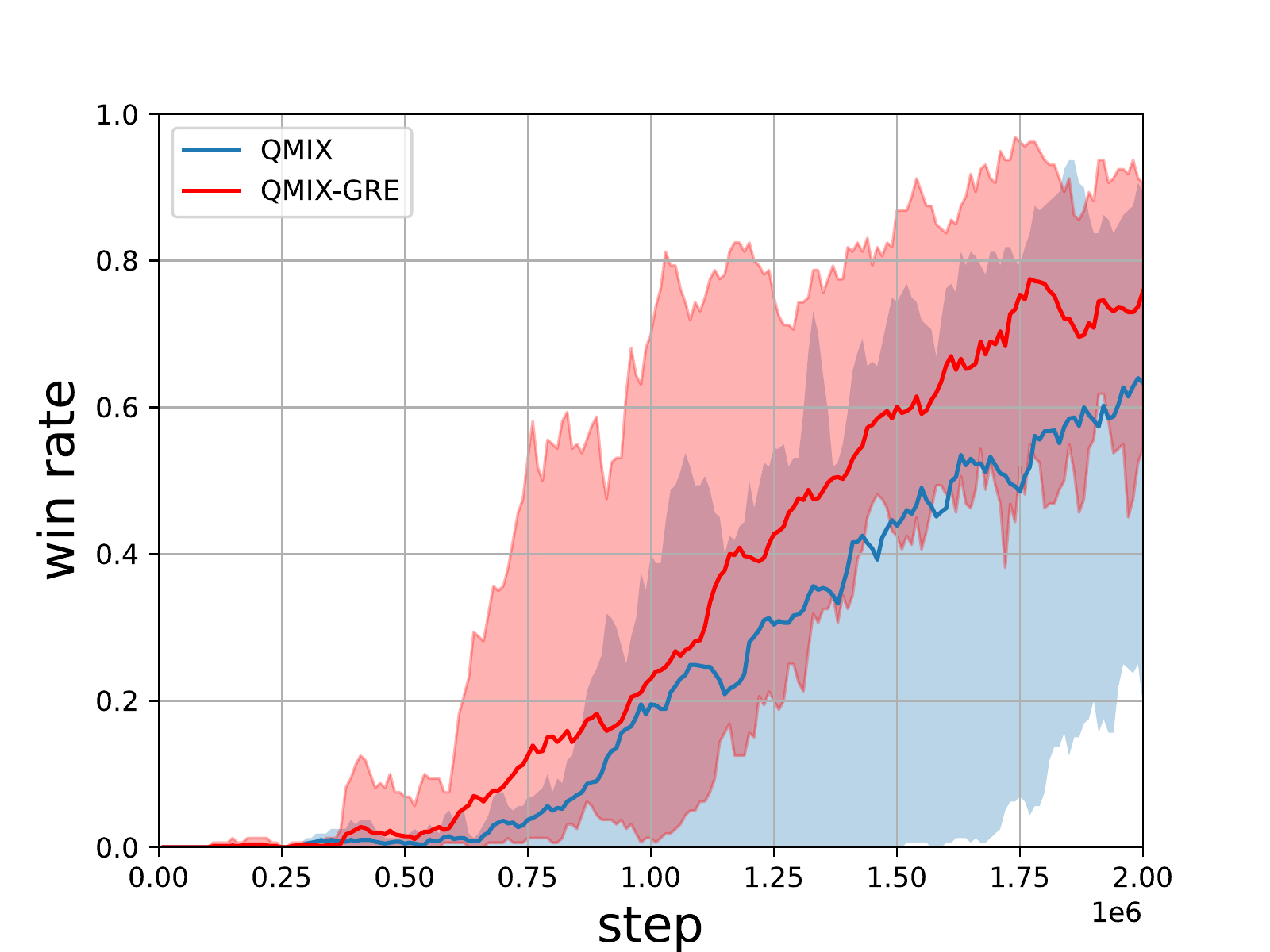}
		\end{minipage}
	}
	\subfigure[27m vs 30m]{
		\begin{minipage}{0.3\textwidth}
			\includegraphics[width=\textwidth]{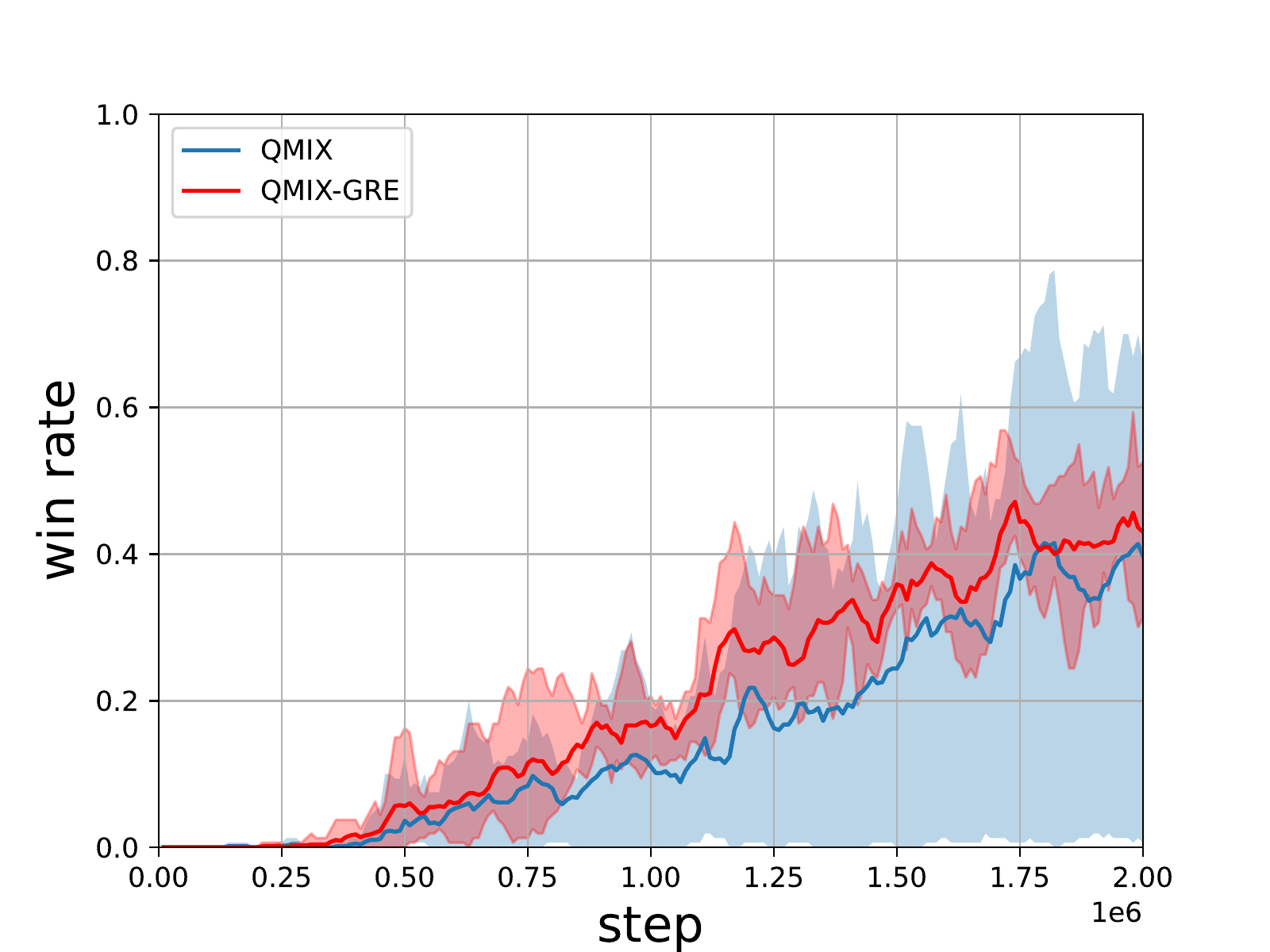}
		\end{minipage}
	}}
	\caption{The performance of QMIX and QMIX-GRE.
	In (a)\emph{Lumberjacks} and (b)\emph{TrafficJunction10} from gridworld, QMIX is able to train a good joint policy, although with a large variance.
	GRE helps the algorithm to more stably achieve a good team policy.
	In (c)\emph{3s5z}, (d)\emph{MMM2} and (e)\emph{27m vs 30m} from \emph{SMAC}, QMIX-GRE speeds up the training and improves the overall performance.} 
	\label{qmix-GRE performance}
\end{figure*}

We test our method on both gridworld and several SMAC maps.
The results are shown in Figure~\ref{qmix-GRE performance}.
More results on SMAC can be found in appendix.
In terms of the overall performance, QMIX-GRE can outperform QMIX or at least achieve comparable performance.
In terms of training efficiency, QMIX-GRE has a significant advantage over QMIX in the early stage, indicating that QMIX-GRE can speed up the training process.

In simple gridworld environments like Lumberjacks and TrafficJunction10, QMIX-GRE can achieve comparable performance.
In such a simple environment, QMIX itself can already solve tasks with high test return, so it is difficult to obtain further improvement.
However, QMIX drops at the middle stage in the training while QMIX-GRE improves the return gradually.
We can see that QMIX-GRE greatly boosts the performance and achieves similar performance on \emph{3s5z}.
On \emph{MMM2} and \emph{27m vs 30m}, QMIX-GRE not only speeds up training, but also improves the overall performance.
The results show that the gradient entropy regularization would be more beneficial to the complex environment.


\begin{figure}[htb]
	\centering
	\subfigure{
		\begin{minipage}{0.35\textwidth} 
            \includegraphics[width=\textwidth]{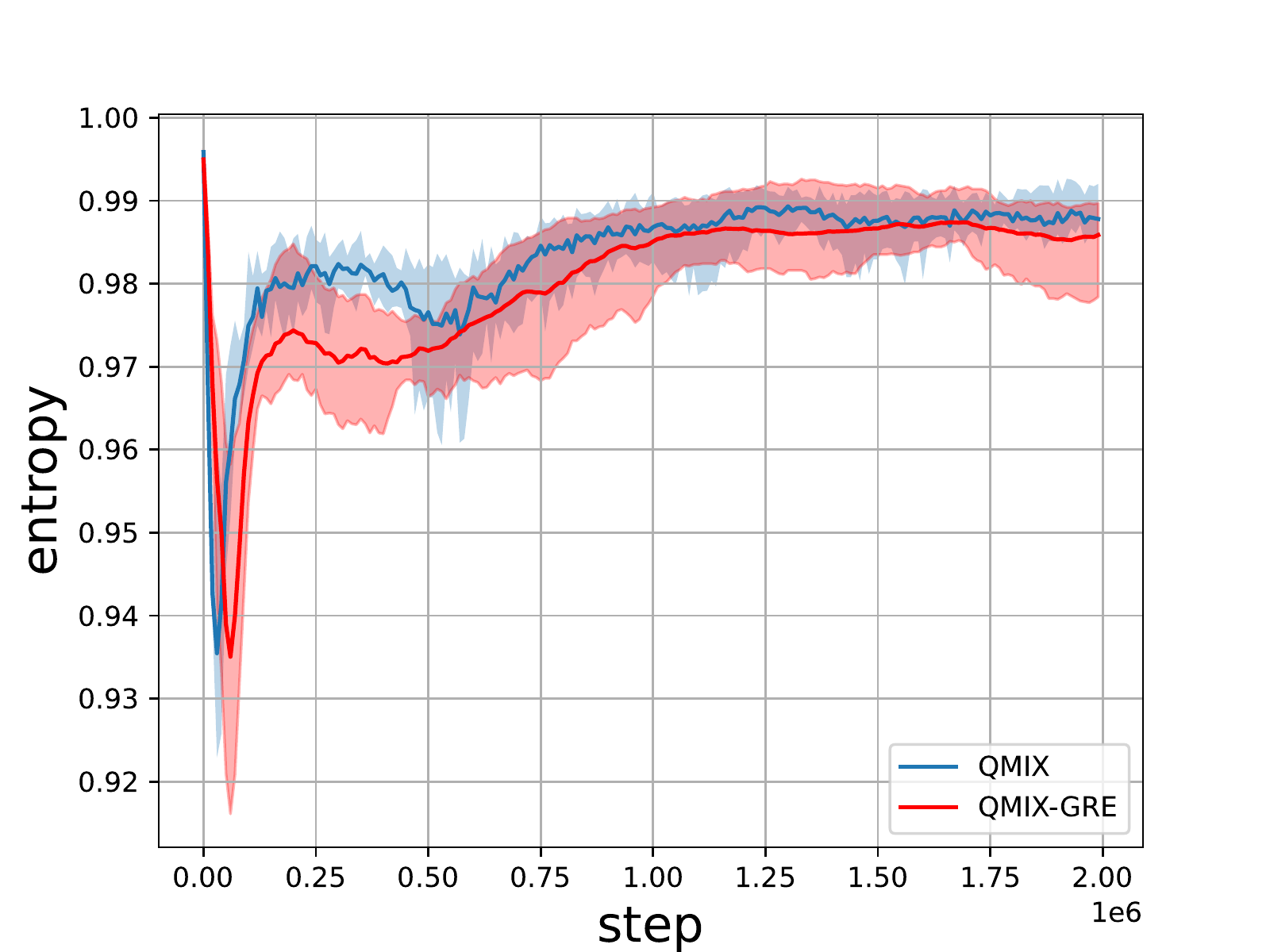}
		\end{minipage}
	}
	\caption{The entropy curve of QMIX and QMIX-GRE on \emph{MMM2} from \emph{SMAC}.
	The normalized entropy of QMIX-GRE significantly decreases, which implies the effectiveness of GRE.} 
	\label{qmix-GRE entropy}
\end{figure}

\subsection{Discriminability of QMIX-GRE: RQ3}
To further investigate whether gradient entropy regularization indeed helps QMIX improve the discriminability of credit assignment, we collect the normalized gradient entropy data in the whole training process and compare it with QMIX.

Due to the page limit, we only plot the entropy curve of \emph{MMM2} in SMAC, as shown in Figure \ref{qmix-GRE entropy}.
We can find that in the random exploration stage in the training, there is actually little difference between the entropy of QMIX and QMIX-GRE.
As training progresses, the entropy of QMIX-GRE is significantly smaller than QMIX, which means that QMIX-GRE tends to realize a more discriminative credit assignment.
It is worth mentioning that the gradient entropy is not that too small, which mainly results from the trade off between the optimization of TD loss and regularization term.

\section{Conclusion}
\label{conclusion}
This paper revisits QMIX and proposes a new measurement to quantitatively study the credit assignment mechanism in algorithms whose framework are similar to QMIX, using the gradient flow and normalized entropy.
With this measurement, we collect numerous data during the training process of QMIX in gridworld environments as well as various SMAC environments.
Our results reveal that QMIX actually does not work as expected in most of the environments to distribute credits differentiably.
To this end, we proposed Gradient entropy REgularization (GRE) to force the mixing network to be more distinguishable.
Experiments show that our method does help boost the training and improve the performance of QMIX.

In our future work, we aim at conducting more experiments on more tasks to verify our normalized entropy measurement and GRE method.
Further, we hope to design more powerful algorithms according to the entropy measurement and investigate deeper into MARL mechanisms.
\newpage

\bibliographystyle{unsrt}  
\bibliography{references}  





\newpage
\appendix
\onecolumn
\section{Appendix}
\subsection{Derivation of $Q_{tot}$ on \{ $Q_i$ \}}
As in section ~\ref{measurement}, we need to calculate the gradients of $Q_{tot}$ on $\{ Q_i \}$.
The detailed process is listed here:

\begin{equation}
\begin{aligned}
    \mathrm{d}Q_{tot}&=\mathrm{d}(W_2f_{elu}(W_1\pmb{Q}+b_1)+b_2) \\
    &=W_2\mathrm{d}f_{elu}(W_1\pmb{Q}+b_1), \\
    \mathrm{d}f_{elu}(\pmb{X})&=\left[ \begin{matrix} 
    \frac{\mathrm{d}f_{elu}(x_1)}{\mathrm{d}x_1} & 0 & \cdots & 0 \\
    0 & \frac{\mathrm{d}f_{elu}(x_2)}{\mathrm{d}x_2} & \cdots & 0 \\
    \vdots & \vdots & \ddots & \vdots \\
    0 & 0 & \cdots & \frac{\mathrm{d}f_{elu}(x_m)}{\mathrm{d}x_m}
    \end{matrix} \right] \mathrm{d}\pmb{X} \\
    \mathrm{d}(W_1\pmb{Q}+b_1)&=W_1\mathrm{d}\pmb{Q},
\end{aligned}
\end{equation}
where $\pmb{Q}=[Q_1(\tau_1, a_1), Q_2(\tau_2, a_2), \cdots, Q_n(\tau_n, a_n)]^T\in \mathbb{R}^{n\times 1}$ is the output of individual Q-networks, and $W_1 \in \mathbb{R}_{+}^{m\times n}$, $W_2 \in \mathbb{R}_{+}^{1 \times m}$, $b_1\in \mathbb{R}^{m \times 1}$ and $b_2\in \mathbb{R}$ are weights produced by hypernetworks.

Therefore, we have
\begin{equation}
    \mathrm{d}Q_{tot}=W_2 F W_1 \mathrm{d}\pmb{Q},
\end{equation}
where $F$ is a diagonal matrix and $F_{ii}=\frac{\mathrm{d}f_{elu}(x_i)}{\mathrm{x_i}}|_{x_i=(W_1\pmb{Q}+b_1)_i}$.

\subsection{Experiments}

We develop our code based on PyMARL framework ~\cite{samvelyan19smac}.
For the environmental settings, we follow the default setting in the QMIX implementation of PyMARL.

\begin{table}[htbp]
\centering
\resizebox{0.5\linewidth}{!} {
\begin{tabular}{c|ccccc}
\hline
\diagbox{Env}{Percentile}          & 5\%     & 25\%    & 50\%    & 75\%    & 95\%    \\ \hline
Lumberjacks       & 0.972 & 0.992 & 0.997 & 0.999 & 1.000 \\
TrafficJunction10 & 0.931 & 0.961 & 0.975 & 0.985 & 0.993 \\
3s vs 5z          & 0.870 & 0.940 & 0.972 & 0.990 & 0.998 \\
3s5z              & 0.941 & 0.975 & 0.984 & 0.989 & 0.995 \\
27m vs 30m        & 0.987 & 0.992 & 0.995 & 0.997 & 0.999 \\
2s3z              & 0.924 & 0.961 & 0.977 & 0.988 & 0.996 \\
5m vs 6m          & 0.975 & 0.988 & 0.993 & 0.996 & 0.999 \\
10m vs 11m        & 0.984 & 0.991 & 0.994 & 0.996 & 0.998 \\
bane vs bane      & 0.917 & 0.970 & 0.983 & 0.990 & 0.995 \\
MMM2              & 0.961 & 0.980 & 0.988 & 0.993 & 0.996 \\ \hline
\end{tabular}}
\caption{The percentiles of the normalized entropy value.
It can be seen that in all of our experiments the $25^{th}$ percentile of gradient entropy is close to 1, and the difference between percentiles is relatively small.
This observation supports our proposal that QMIX actually fail to assign credits discriminately.}
\end{table}

We first calculate the entropy during the whole training process in two gridworld environments and several SMAC environments.
The $q^{th}$ ($q \in \{5,25,50,75,95\}$) percentile of entropy value for each map/environment is shown below, while some of them have already been shown in the main body of this paper. 
The data is collected in the whole training process, where in gridworld we store the gradients every 1K steps and 200K steps in total and in SMAC we save the gradients every 10K steps and 2M steps in total.
The percentiles are calculated with all the saved data.

\begin{figure}[htbp]
	\centering
	\subfigure[Lumberjacks]{
		\begin{minipage}{0.2\textwidth}
			\includegraphics[width=\textwidth]{imgs/grid_world/entropy/Lumber_entropy.pdf}
		\end{minipage}
	}
	\subfigure[TrafficJunction10]{
		\begin{minipage}{0.2\textwidth}
			\includegraphics[width=\textwidth]{imgs/grid_world/entropy/Traffic10_entropy.pdf}
		\end{minipage}
	}
	\\
	\subfigure[3s vs 5z]{
		\begin{minipage}{0.2\textwidth} 
            \includegraphics[width=\textwidth]{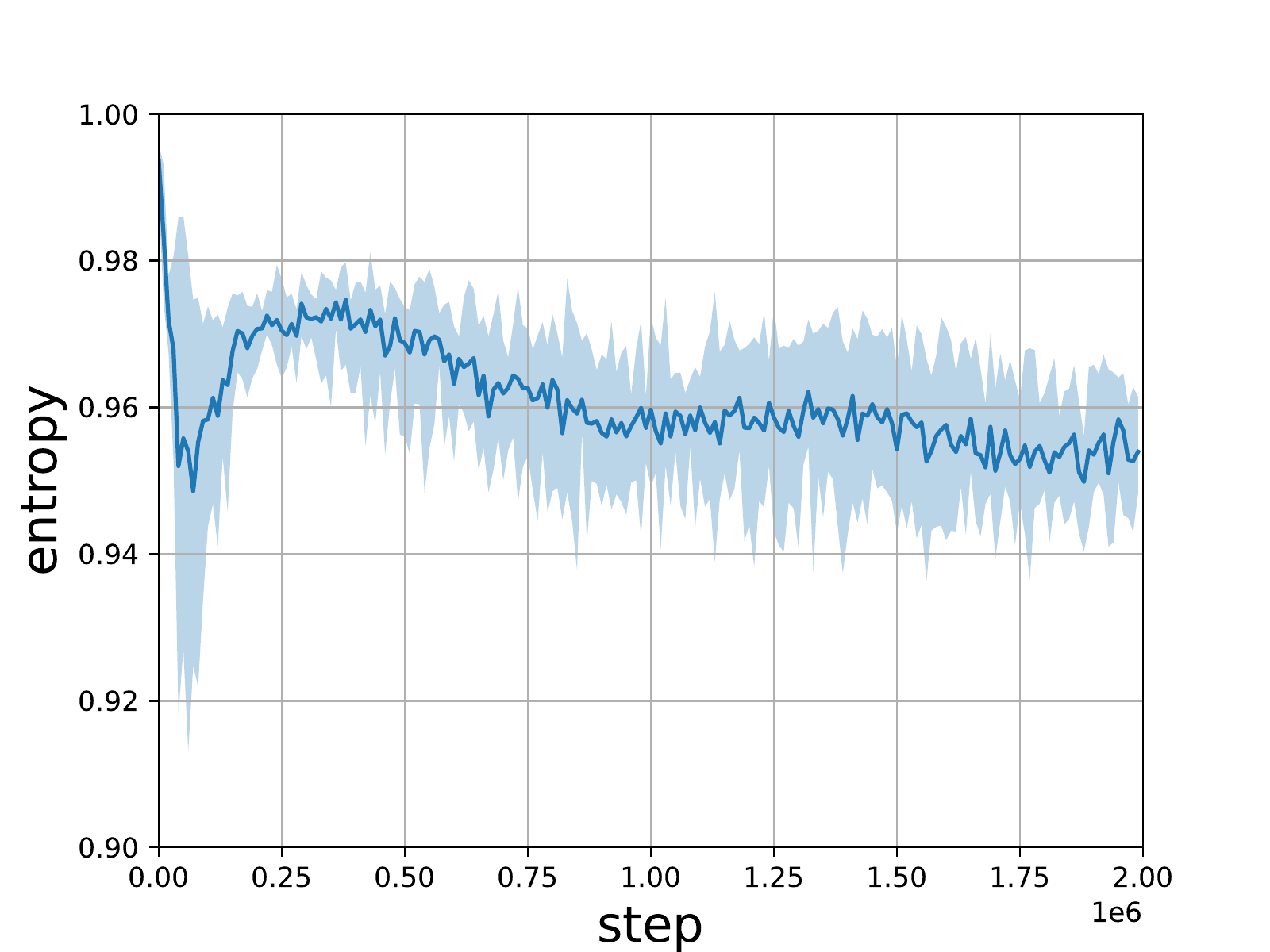}
		\end{minipage}
	}
	\subfigure[3s5z]{
		\begin{minipage}{0.2\textwidth}
			\includegraphics[width=\textwidth]{imgs/smac/entropy_curve/3s5z_curve.pdf}
		\end{minipage}
	}
	\subfigure[27m vs 30m]{
		\begin{minipage}{0.2\textwidth}
			\includegraphics[width=\textwidth]{imgs/smac/entropy_curve/27m_vs_30m_curve.pdf}
		\end{minipage}
	}
	\subfigure[2s3z]{
		\begin{minipage}{0.2\textwidth} 
            \includegraphics[width=\textwidth]{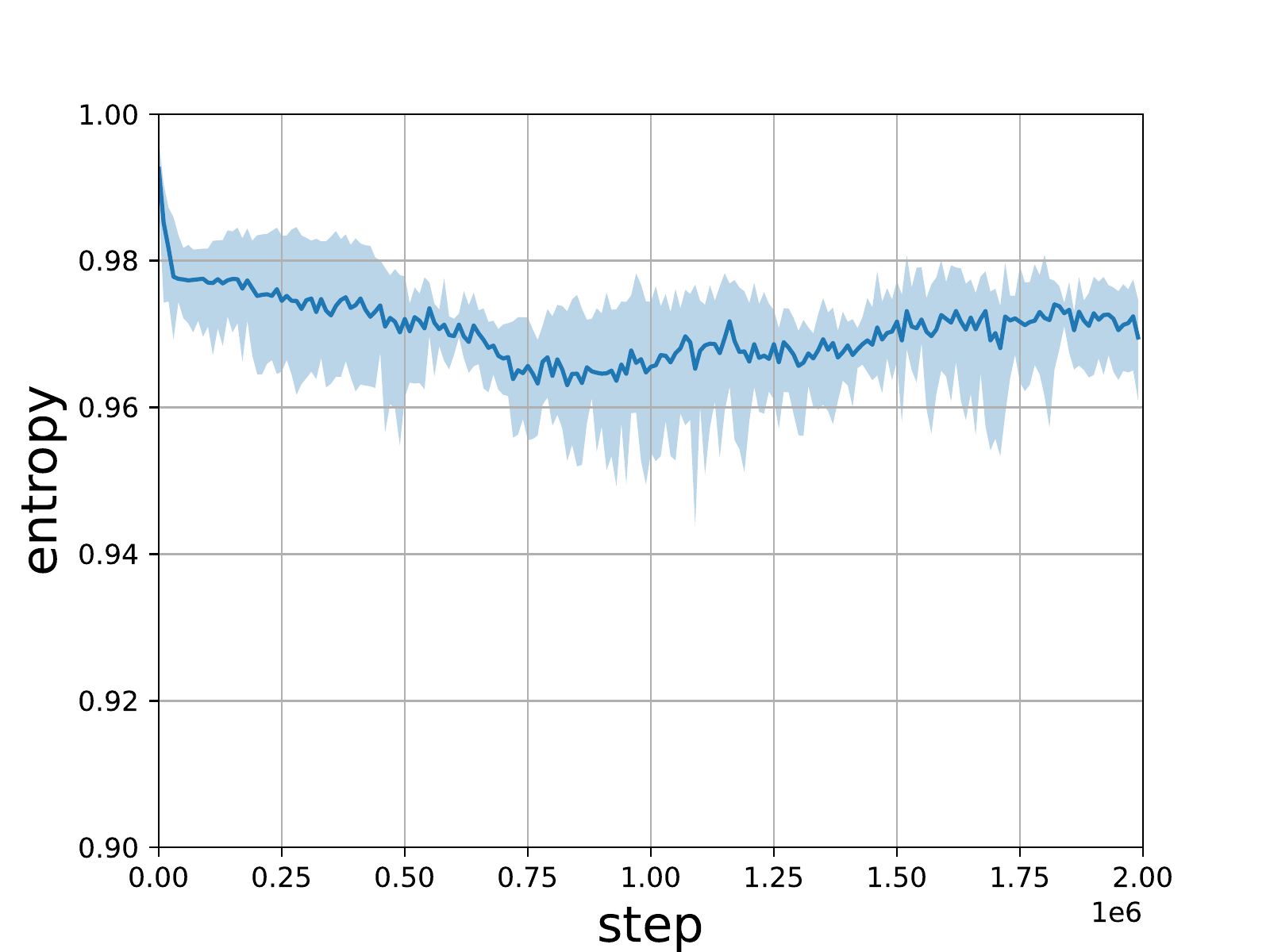}
		\end{minipage}
	}
	\\
	\subfigure[5m vs 6m]{
		\begin{minipage}{0.2\textwidth}
			\includegraphics[width=\textwidth]{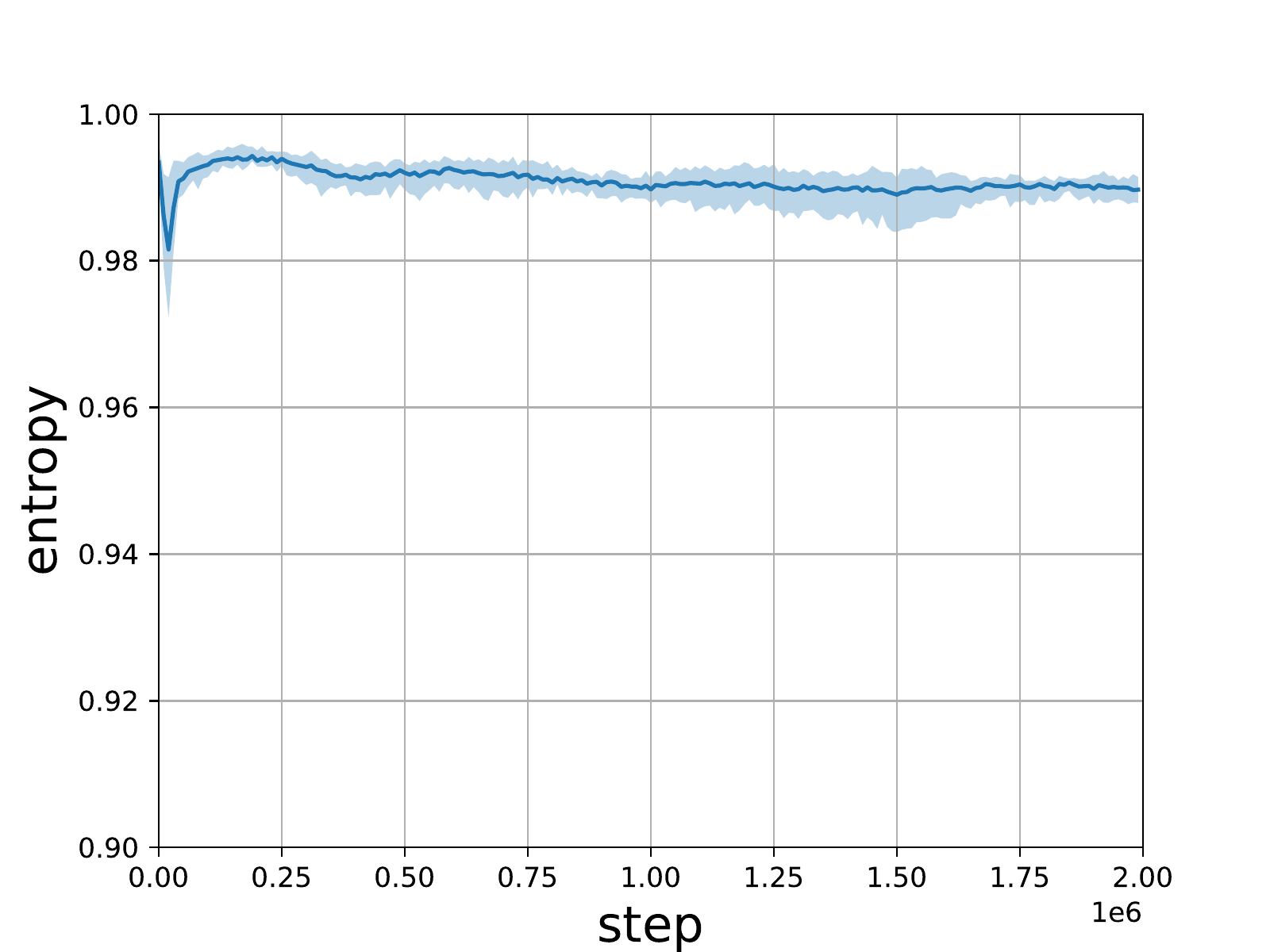}
		\end{minipage}
	}
	\subfigure[10m vs 11m]{
		\begin{minipage}{0.2\textwidth}
			\includegraphics[width=\textwidth]{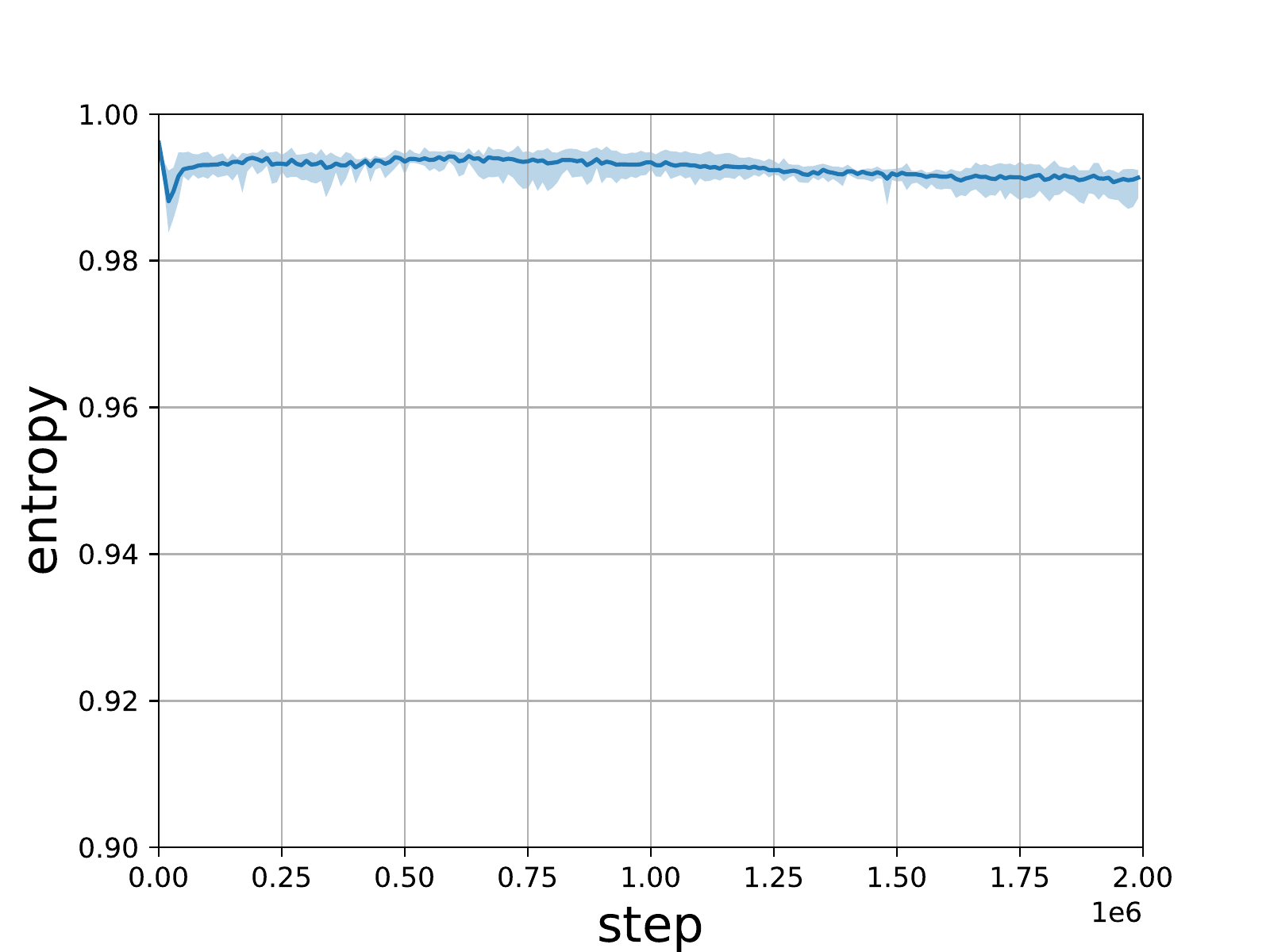}
		\end{minipage}
	}
	\subfigure[bane vs bane]{
		\begin{minipage}{0.2\textwidth}
			\includegraphics[width=\textwidth]{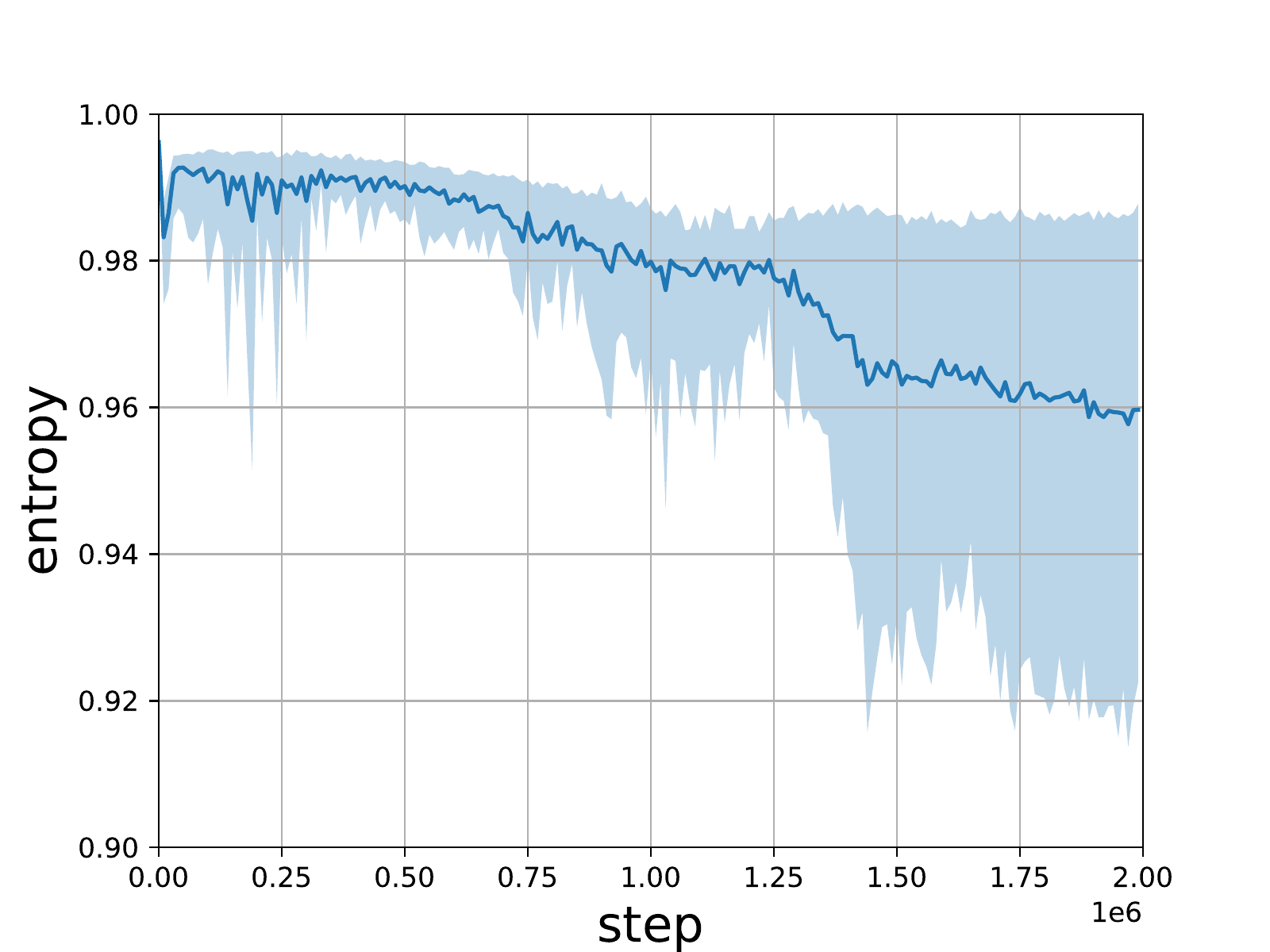}
		\end{minipage}
	}
	\subfigure[MMM2]{
		\begin{minipage}{0.2\textwidth}
			\includegraphics[width=\textwidth]{imgs/smac/entropy_curve/MMM2_curve.pdf}
		\end{minipage}
	}
	\caption{The entropy curve of QMIX. (ranging from 0.9 to 1.0)
	Both of the solid line are the mean performance from 5 different random seeds.
	The upper and lower bound of the shadow area are min and max values in the 5 seeds.
	}
	\label{qmix_entropy_all_env}
\end{figure}

And Figure~\ref{qmix_entropy_all_env} shows the normalized entropy in the training process on these environments.

\begin{figure}[htbp]
	\centering
	\subfigure[Lumberjacks]{
		\begin{minipage}{0.2\textwidth}
			\includegraphics[width=\textwidth]{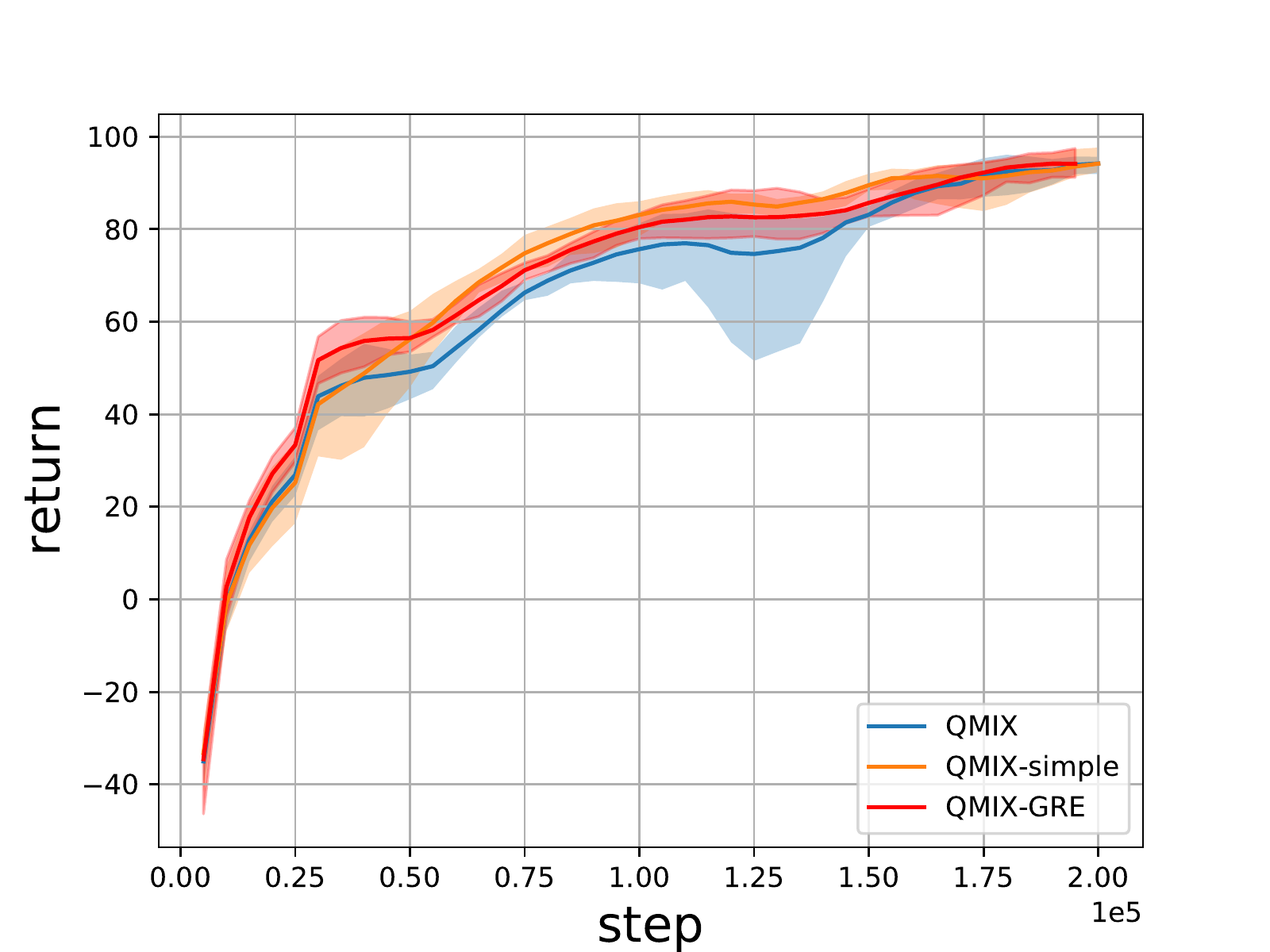}
		\end{minipage}
	}
	\subfigure[TrafficJunction10]{
		\begin{minipage}{0.2\textwidth}
			\includegraphics[width=\textwidth]{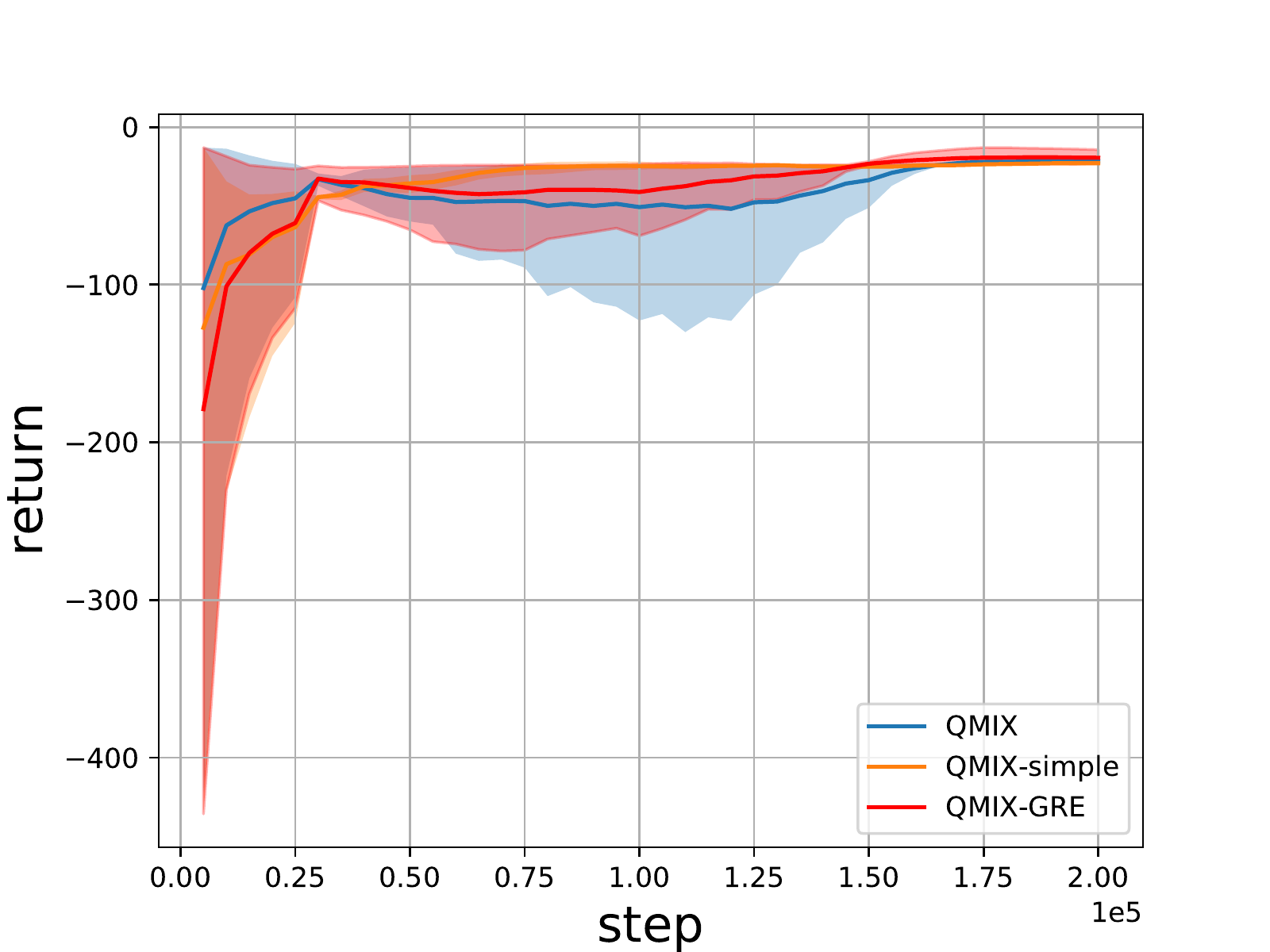}
		\end{minipage}
	}
	\\
	\subfigure[3s vs 5z]{
		\begin{minipage}{0.2\textwidth} 
            \includegraphics[width=\textwidth]{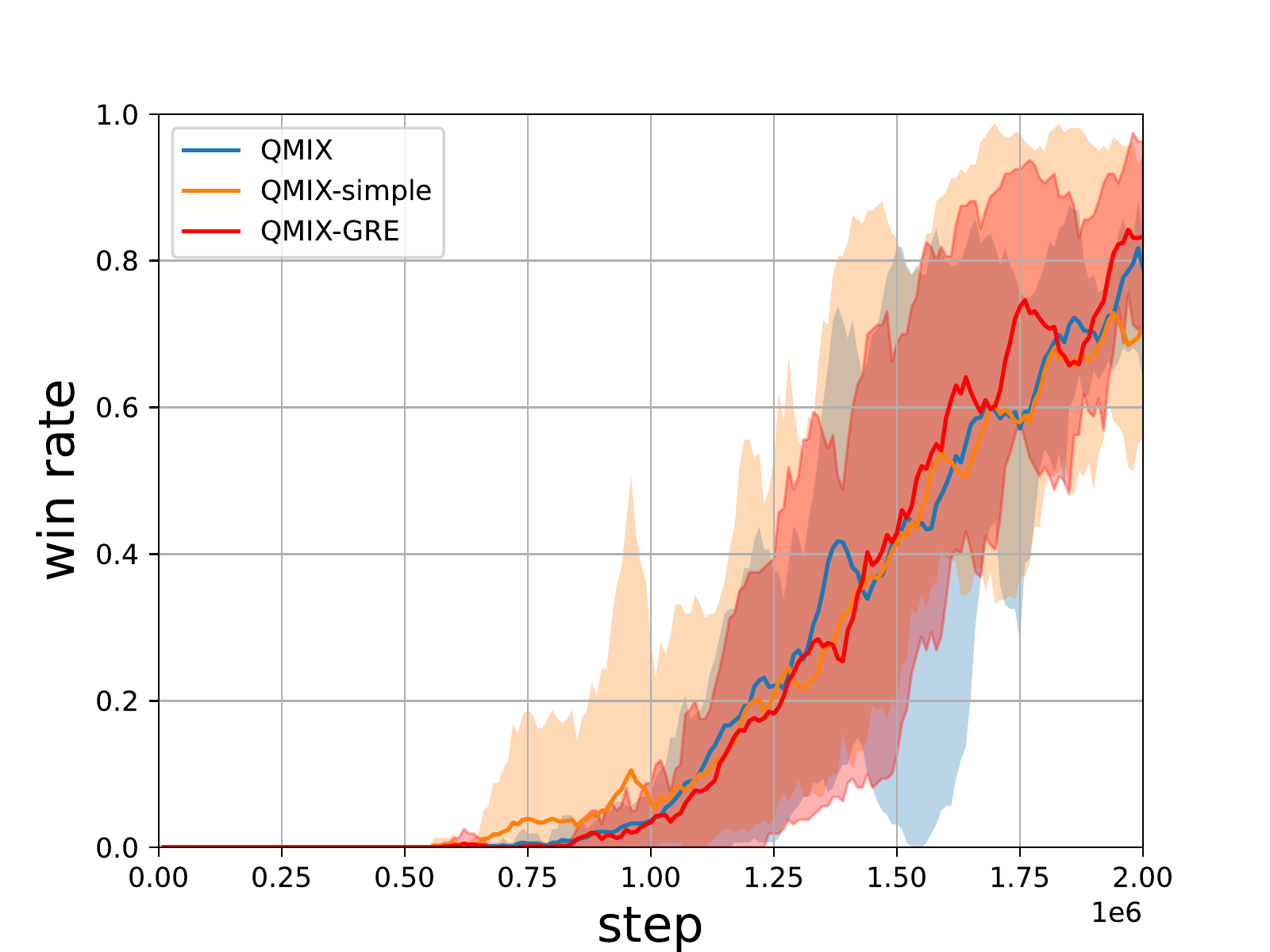}
		\end{minipage}
	}
	\subfigure[3s5z]{
		\begin{minipage}{0.2\textwidth}
			\includegraphics[width=\textwidth]{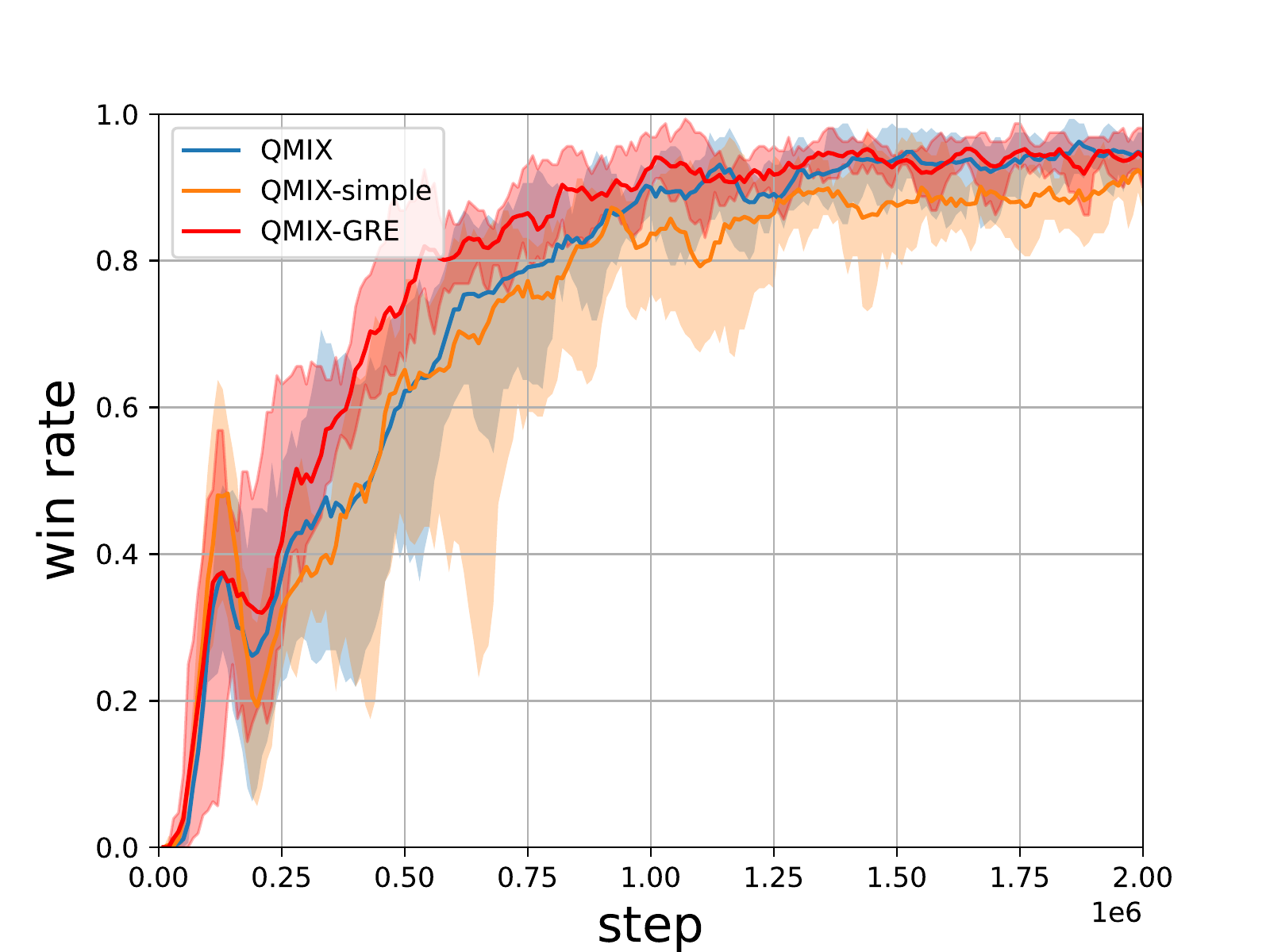}
		\end{minipage}
	}
	\subfigure[27m vs 30m]{
		\begin{minipage}{0.2\textwidth}
			\includegraphics[width=\textwidth]{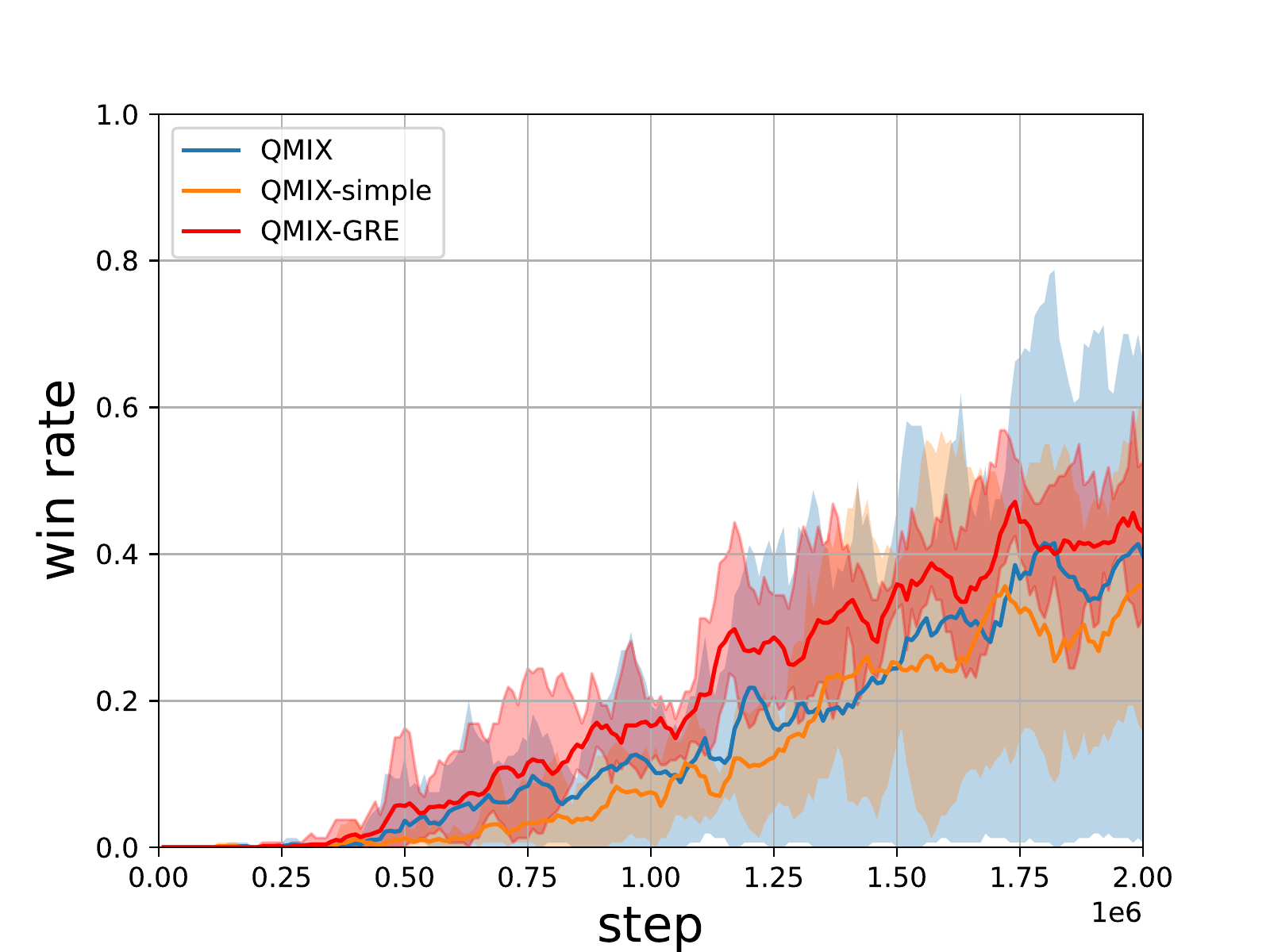}
		\end{minipage}
	}
	\subfigure[2s3z]{
		\begin{minipage}{0.2\textwidth} 
            \includegraphics[width=\textwidth]{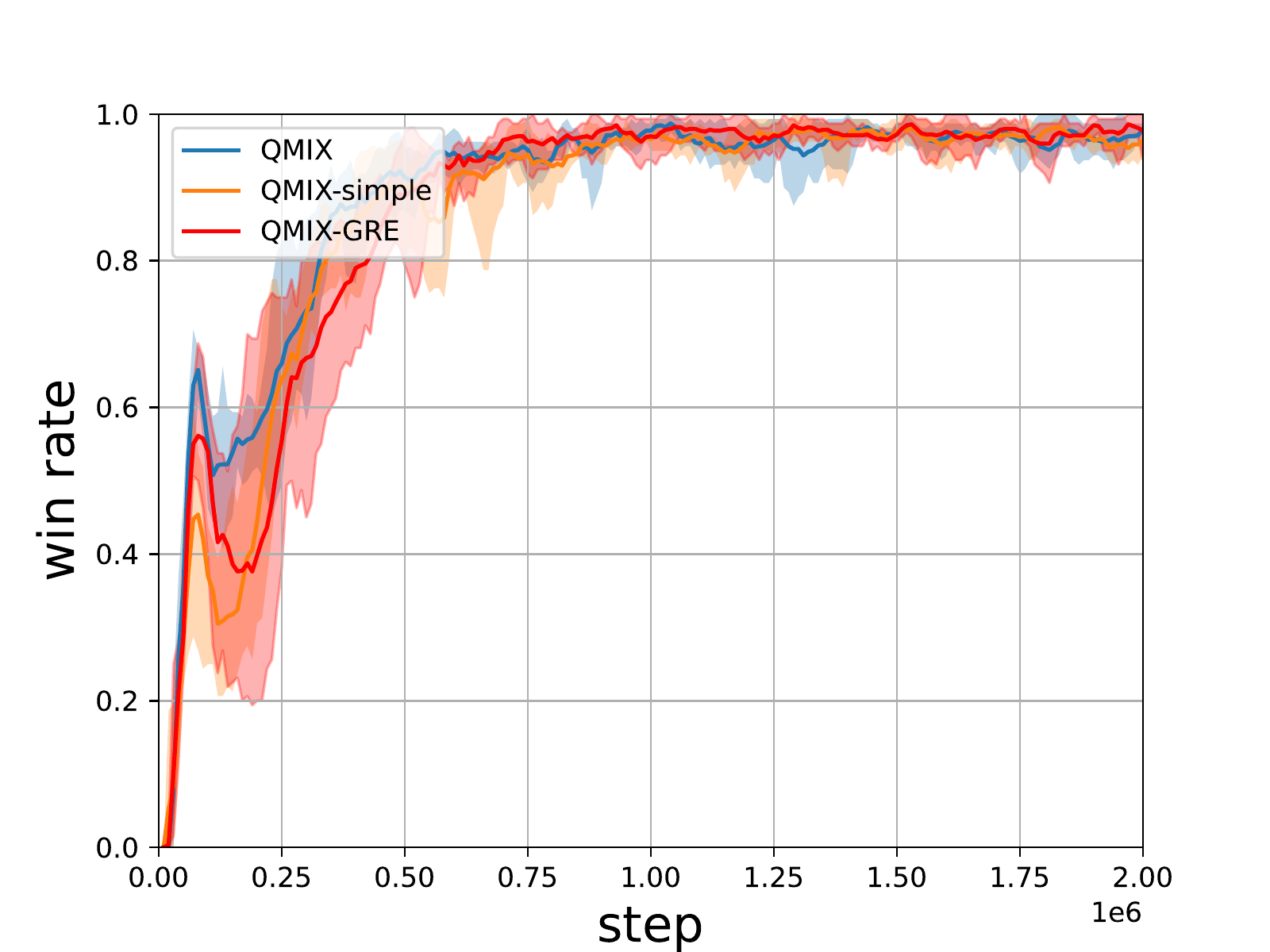}
		\end{minipage}
	}
	\\
	\subfigure[5m vs 6m]{
		\begin{minipage}{0.2\textwidth}
			\includegraphics[width=\textwidth]{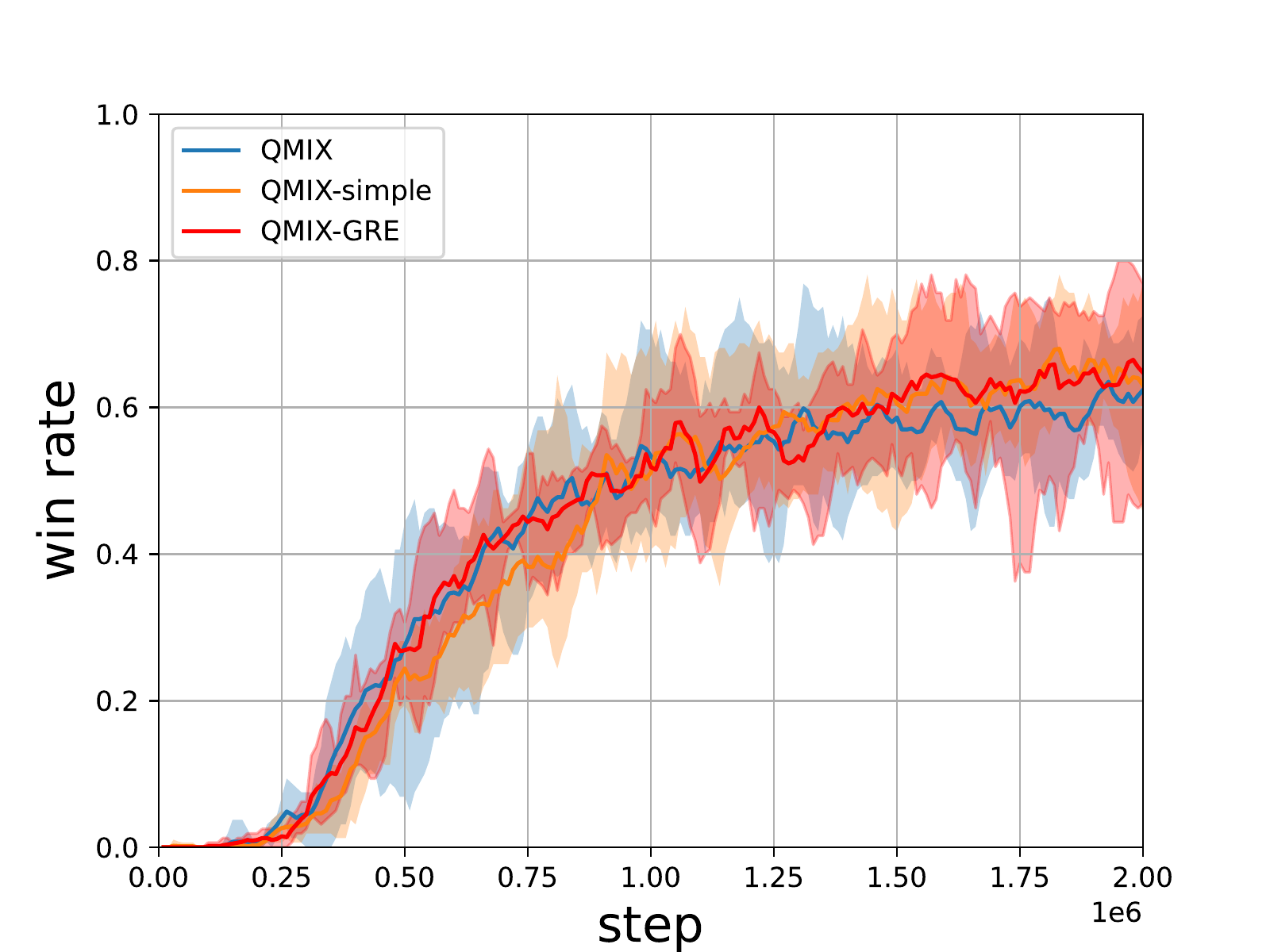}
		\end{minipage}
	}
	\subfigure[10m vs 11m]{
		\begin{minipage}{0.2\textwidth}
			\includegraphics[width=\textwidth]{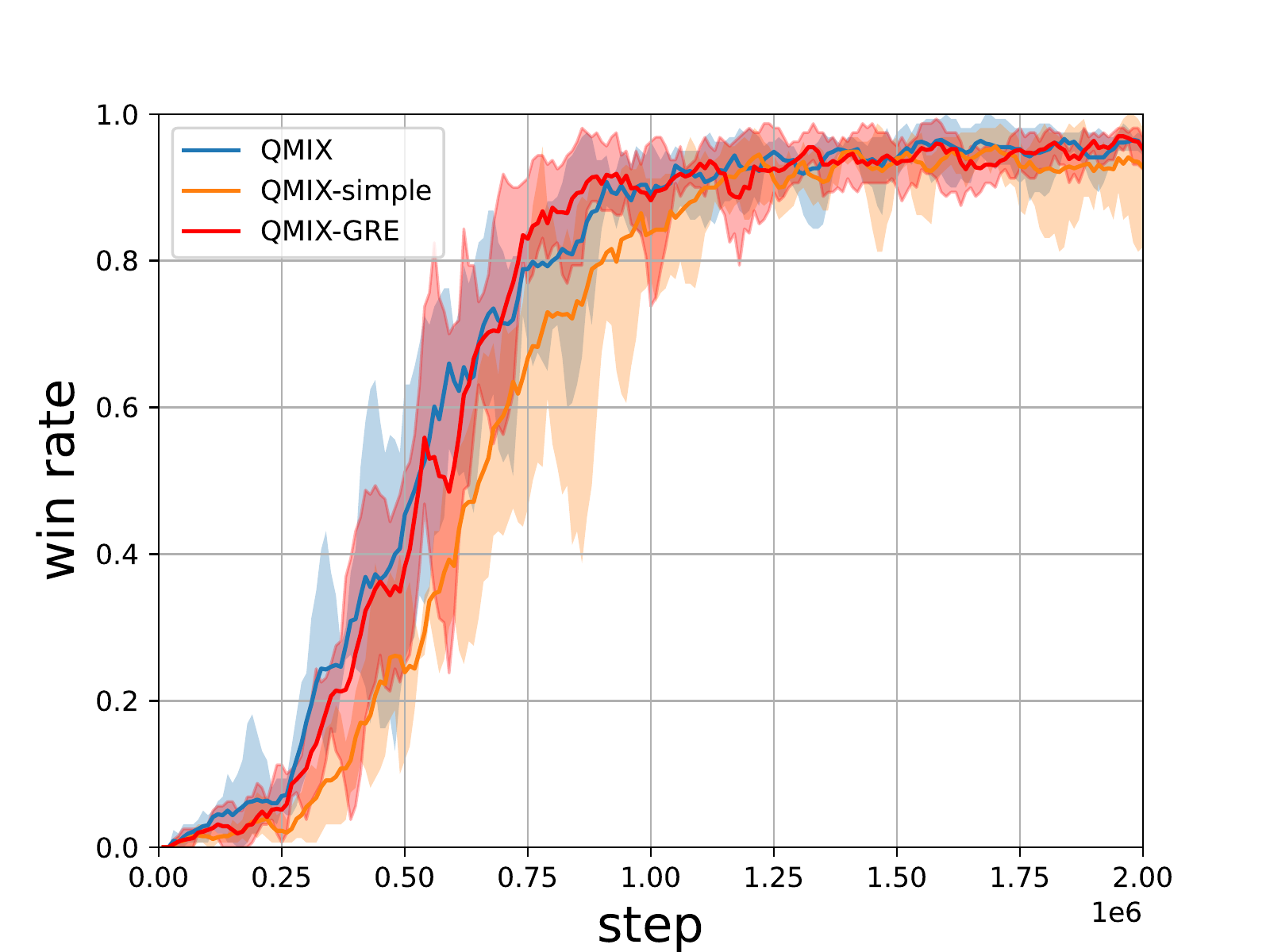}
		\end{minipage}
	}
	\subfigure[bane vs bane]{
		\begin{minipage}{0.2\textwidth}
			\includegraphics[width=\textwidth]{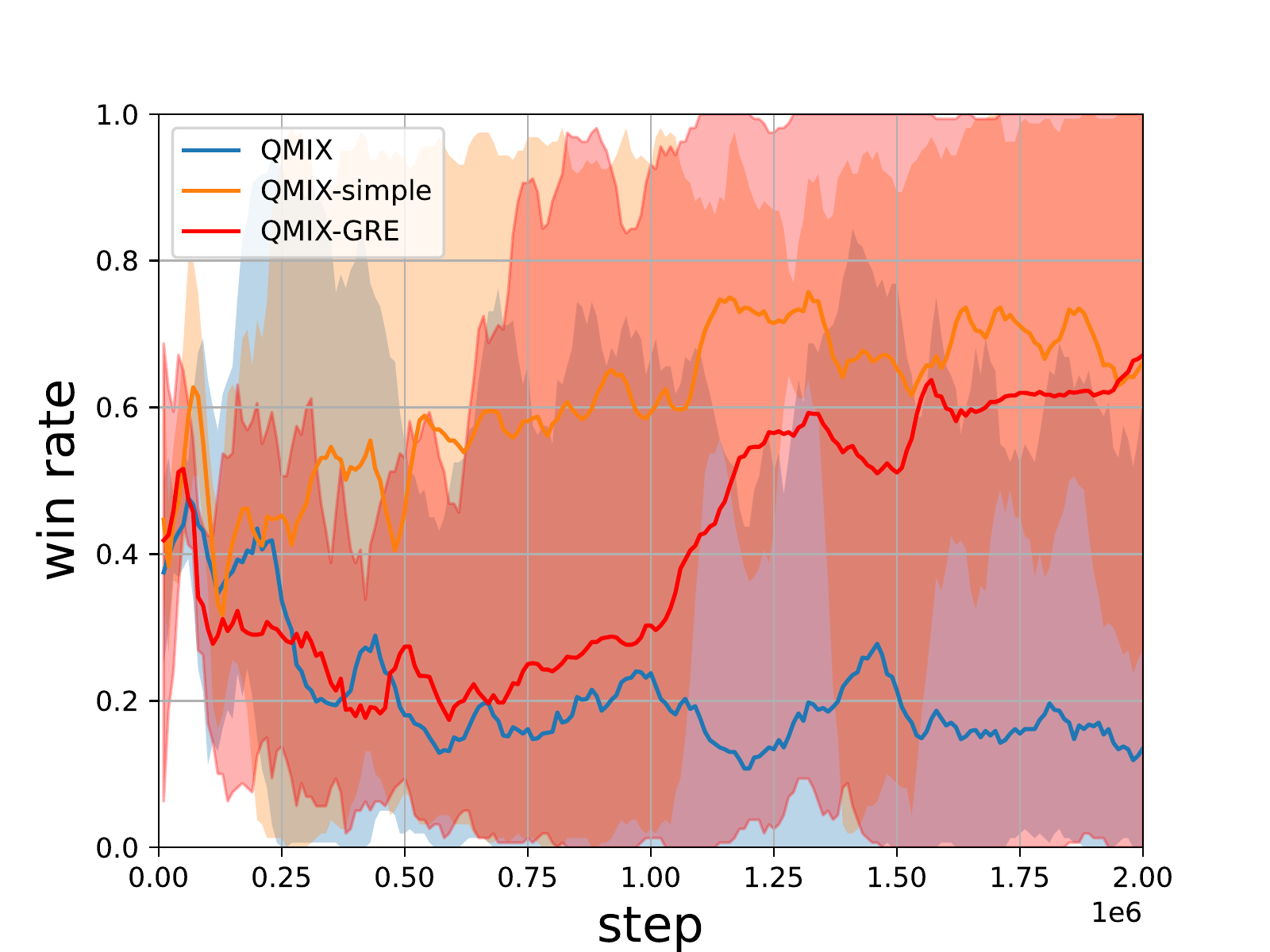}
		\end{minipage}
	}
	\subfigure[MMM2]{
		\begin{minipage}{0.2\textwidth}
			\includegraphics[width=\textwidth]{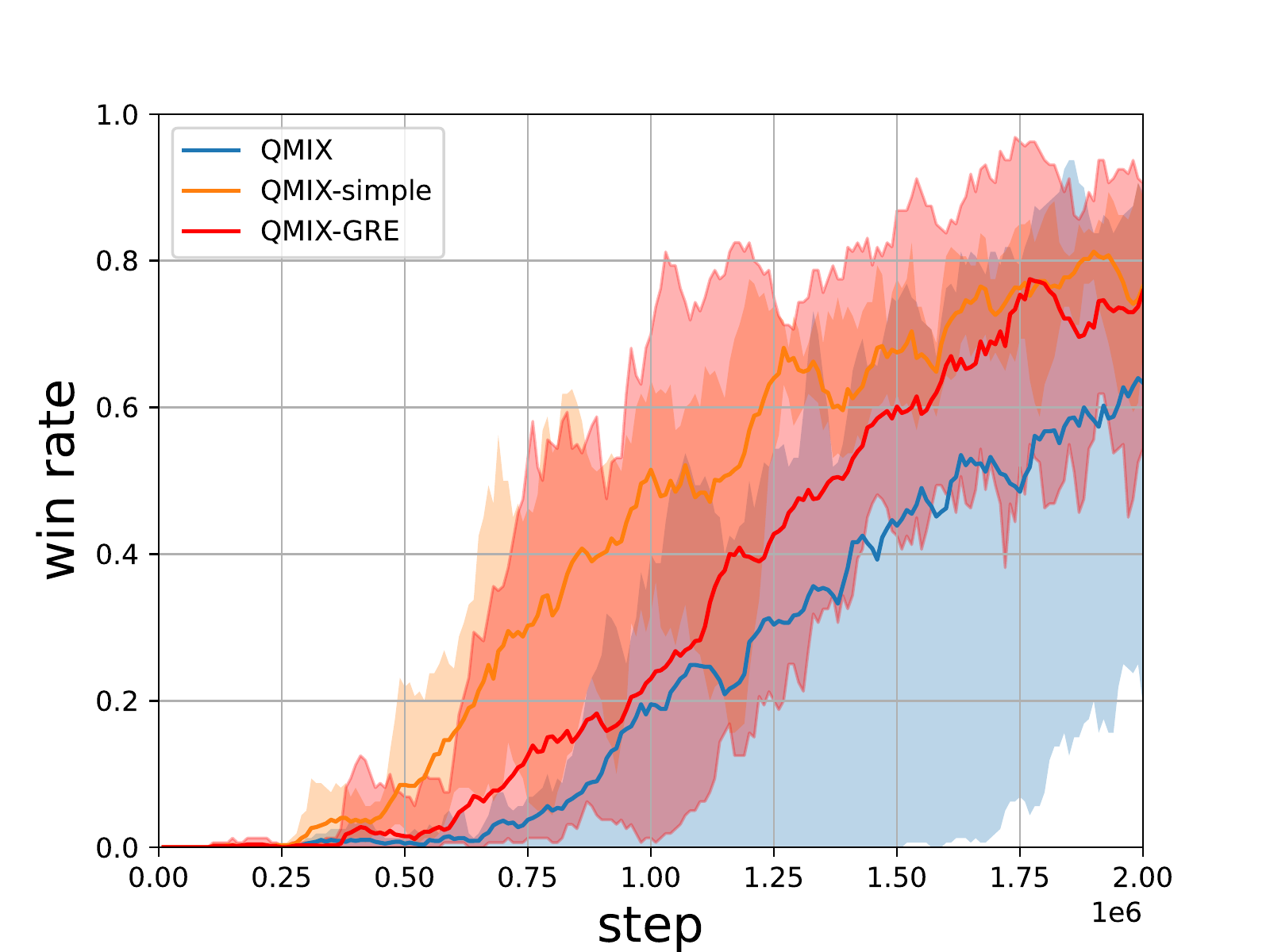}
		\end{minipage}
	}
	\caption{The performance of QMIX, QMIX-simple and QMIX-GRE.
	Both of the solid line are the mean performance from 5 different random seeds.
	The upper and lower bound of the shadow area are min and max values in the 5 seeds. 
	The $\lambda$ is chosen with grid-search from \{ 1e-3, 5e-4, 1e-4 \}.
	In the future, we will conduct more experiments on the choice of $\lambda$ and dive deeper with this algorithm.
	} 
	\label{qsimple_performance_all_env}
\end{figure}

We also conduct QMIX-simple experiments on these environments, and the results are shown in Figure ~\ref{qsimple_performance_all_env}, some of which have been shown in the main body of the paper.

\end{document}